\documentclass[runningheads]{llncs}

\usepackage{eccv}

\usepackage{eccvabbrv}
\usepackage{svg}
\usepackage{amsmath}
\usepackage{graphicx}
\usepackage{booktabs}
\usepackage{wrapfig}

\usepackage{algpseudocode}
\usepackage[ruled,vlined]{algorithm2e}

\usepackage[accsupp]{axessibility}  

\usepackage{hyperref}
\usepackage{xcolor}
\definecolor{lightblue}{RGB}{43,127,255} 

\hypersetup{
    colorlinks=true,
    urlcolor=lightblue
}

\usepackage{orcidlink}

\begin{document}

\title{Beyond Prompts: Unconditional 3D Inversion for Out-of-Distribution Shapes} 

\titlerunning{Beyond Prompts}

\author{Victoria Yue Chen \inst{1}\orcidlink{0009-0000-3353-3768} \and
Emery Pierson \inst{2}\orcidlink{0009-0006-5243-2974} \and
Léopold Maillard \inst{2,3}\orcidlink{0009-0002-8974-0648} \and
Maks~Ovsjanikov \inst{2}\orcidlink{0000-0002-5867-4046}
}

\authorrunning{V.~Chen et al.}

\institute{ETH Zürich \and
LIX, École Polytechnique, IP Paris \and
Dassault Systèmes 
}

\maketitle

\begin{abstract}
Text-driven inversion of generative models is a core paradigm for manipulating 2D or 3D content, unlocking numerous applications such as text-based editing, style transfer, or inverse problems. However, it relies on the assumption that generative models remain sensitive to natural language prompts. We demonstrate that for state-of-the-art native text-to-3D generative models, this assumption often collapses. We identify a critical failure mode where generation trajectories are drawn into latent ``sink traps'': regions where the model becomes insensitive to prompt modifications. In these regimes, changes to the input text fail to alter internal representations in a way that alters the output geometry. Crucially, we observe that this is not a limitation of the model's \textit{geometric} expressivity; the same generative models possess the ability to produce a vast diversity of shapes but, as we demonstrate, become insensitive to out-of-distribution \textit{text} guidance. We investigate this behavior by analyzing the sampling trajectories of the generative model, and find that complex geometries can still be represented and produced by leveraging the model's unconditional generative prior. This leads to a more robust framework for text-based 3D shape editing that bypasses latent sinks by decoupling a model's geometric representation power from its linguistic sensitivity. Our approach addresses the limitations of current 3D pipelines and enables high-fidelity semantic manipulation of out-of-distribution 3D shapes. Project webpage: \url{https://daidedou.sorpi.fr/publication/beyondprompts}
\end{abstract}

\section{Introduction}
\label{sec:intro}

Inverting generative models consists of mapping a given input into the noise space of a pre-trained network. In recent years, this approach has been shown to enable many analysis and editing applications thanks to its flexibility and efficiency (\eg, \cite{nti,wang2024taming2} among myriad others). The process typically relies on a paired visual input (\eg, an image or a 3D shape) and an approximate text prompt, representing this input. By embedding the source (\eg, ``rabbit'') in the model noise space, text-based editing and manipulation can be done by simply modifying the prompt (\eg, ``rabbit lunging forward''), followed by denoising to obtain the edited asset. This simplicity, where the user only has to modify a text prompt instead of relying on complex, specialized software, can be leveraged in many scenarios such as text-driven editing~\cite{nti}, style transfer~\cite{zhang2023inversion}, or image inverse problems~\cite{chihaoui2024blind}.

Despite the appeal of this strategy, we observe that current 3D manipulation pipelines do not \textit{exclusively} use a 3D generative model or follow the simple strategy described above. 
Instead, to achieve plausible results, they often require auxiliary guidance such as user-defined masks, 2D in-painting models, or exploit strong image-based priors~\cite{parelli20253dlattelatentspace3d, ye2025nano3d, li2025voxhammer}. \textit{Exclusively} using text prompts has not yet been shown to output high-quality inversion or editing results of native text-to-3D generative models. 

\begin{figure}[t]
  \centering
   \includegraphics[width=0.8\linewidth]{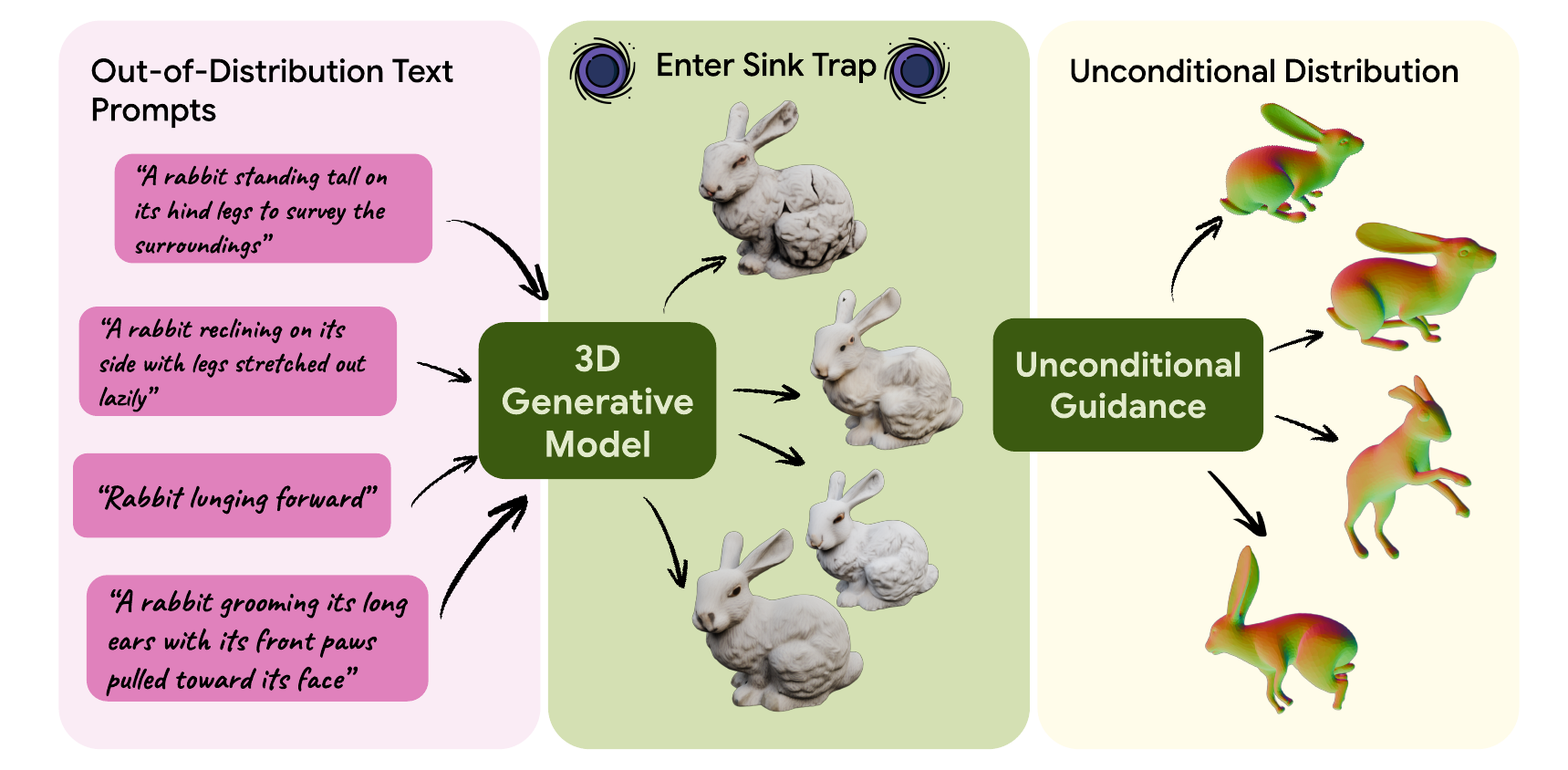}
   \caption{Geometry vs. Language Diversity. (Left) Text-conditioned generation exhibits a ``sink trap'' effect, where diverse prompts for a ``rabbit'' yield nearly identical geometries. (Right) In contrast, our unconditioned 3D generative model overcomes this linguistic bottleneck, faithfully inverting and reconstructing arbitrary 3D shapes with high fidelity.}
   \label{fig:stanford_bunny}
\end{figure}

In this paper we analyze the root causes behind this phenomenon and reveal a \textit{fundamental mismatch} between the \textit{geometric} expressivity of text-to-3D generative models and their \textit{text}-reasoning capacity. Specifically, we note that, as illustrated in Fig.~\ref{fig:stanford_bunny}, edit failures often do not stem from the generative model’s inability to represent a target \textit{geometry}, but rather from the prompt falling out-of-distribution. We identify this failure mode as the ``sink trap'': a state where the model becomes unresponsive to textual variations, consistently producing a nearly identical shape regardless of the prompt, despite possessing the capacity to produce a variety of geometric structures.

We analyze this behavior by investigating the internal sampling trajectories of 3D generative models, and observe that they become highly unstable when using mismatched or out-of-distribution prompts during inversion. Based on this analysis, we find that leveraging the \textit{unconditional distribution} of the generative model is crucial to producing stable trajectories. Building on these insights, we propose an approach that enables the faithful inversion of complex, non-rigid shapes that fail with the standard null-text inversion with approximate prompts \cite{nti} in the context of text-to-3D generative models. Remarkably, despite using the unconditional prior during inversion, we show that our approach still enables \textit{3D shape editing}, driven purely by the target text prompt without the need for any auxiliary models or image-based priors. To the best of our knowledge, ours is the first approach that demonstrates high-fidelity 3D editing using \textit{only a} native text-to-3D generative model and without any additional input beyond the target prompt. Our contributions are summarized as follows:
\begin{enumerate}
\item Characterization of the ``sink trap'' phenomenon: We identify and analyze a specific failure mode in 3D generative models where textual conditioning fails to trigger geometric changes, showing that these failures often stem from a mismatch between a model's text reasoning and geometric representational capabilities. 
\item Trajectory Stabilization via Unconditional Priors: We provide a study of the internal sampling trajectories of 3D models and introduce a method to stabilize them by leveraging the unconditional distribution. This enables the reconstruction of complex, non-rigid shapes where standard text-conditioning cannot succeed.
\item A Text-to-3D Editing Framework: We show that our proposed inversion unlocks the first high-fidelity 3D editing pipeline driven solely by a target text prompt. Our approach relies solely on a native 3D generative model, removing the need for auxiliary image-based priors.
\end{enumerate}

\section{Related Work}
\label{sec:related_work}

\subsection{Image Based Inversion}
Inversion is a prerequisite for editing real-world images with Text-to-Image (T2I) models, as it maps a data sample back to a latent code within the model's noise space by reversing the sampling process. This conversion provides a representation that the model can manipulate, a process fundamental to text-driven editing with diffusion models \cite{song2019generative, ddpm, ddim}. However, direct DDIM inversion often yields imperfect reconstructions due to numerical error accumulation \cite{edict, nti} and the challenges of classifier-free guidance (CFG) \cite{cfg}. While various methods address these inaccuracies (see \cite{Huang_2025} for a survey), Null-Text Inversion (NTI) \cite{nti} remains a cornerstone approach. By optimizing unconditional embeddings, NTI ensures that the edited result maintains high fidelity to the original image while remaining responsive to new textual prompts. 

Similar principles have recently been extended to flow-based generative models \cite{lipman2022flow, liu2022flow}, including \textit{inversion-free} frameworks \cite{kulikov2025flowedit, jiao2025uniedit, xu2025unveil} that bypass iterative optimization and mapping to the noise space. Notably, these inversion-free methods still require mapping the target into a latent representation and necessitate a source prompt that describes the original content. As observed in NTI \cite{nti}, any prompt that is ``reasonable and editable'' would work. This requirement persists when extending 2D solutions to 3D: both the source shape and a corresponding text prompt are required to anchor the asset within the model's latent space for subsequent inversion or editing. Despite the prominence of these methods in the image domain, the applicability of these latent-space manipulation techniques to the native latent spaces of 3D generative models remains largely unexamined.

\subsection{3D content manipulation with Generative Models}
Common approaches to 3D asset manipulation can be categorized into three paradigms: (1) Score Distillation Sampling (SDS)~\cite{poole2022dreamfusion}, which optimizes assets via image generative priors; (2) multiview-consistent image generation, which leverages multiview priors to ensure geometric coherence; and (3) direct manipulation of 3D representations coupled with 2D inpainting models. 

Early work in 3D editing primarily relied on distilling gradients from pre-trained 2D diffusion models~\cite{sella2023voxetextguidedvoxelediting, kim2025meshup, dinh2025geometry}. By rendering 3D assets from stochastic viewpoints, a generative model can output a score that guides optimization toward a text-aligned result. While applicable to both Neural Radiance Fields (NeRF) and explicit meshes, this per-edit optimization process is computationally expensive and slow.

Consequently, many subsequent works leverage multiview image generation for 3D editing~\cite{erkocc2025preditor3d, long2024wonder3d, chen2023control3d, gao20253d, li2025cmd}. These methods are generally more efficient, relying on multiview-consistent diffusion models to generate edited views, followed by a rapid reconstruction or refinement step to ensure global geometric consistency.

Finally, emerging works attempt to leverage state-of-the-art 3D models~\cite{xiang2024structured} directly for text-based editing~\cite{li2025voxhammer, ye2025nano3d}. Notably,~\cite{ye2025nano3d} employs Gemini 2.5 Flash~\cite{gemini25} for inpainting, feeding the edited result back into~\cite{xiang2024structured} for asset generation, while VoxHammer~\cite{li2025voxhammer} utilizes Flux~\cite{labs2025flux1} for inpainting rendered assets. Rather than manipulating internal model latents, these works introduce the editing signal by modifying the conditioning input. Their primary contribution lies in leveraging internal voxel representations to ensure the edited asset maintains structural fidelity to the original. More recently, \cite{parelli20253dlattelatentspace3d} used inversion in the multi-view space. However, due to instabilities in the inversion process, they rely heavily on 2D backbones for mask selection and multi-view optimization of the geometry.

In summary, existing 3D editing frameworks rarely rely exclusively on the native generative capabilities of 3D models, instead incorporating 2D text-to-image (T2I) priors to ensure high visual fidelity. Our study isolates the 3D generative model to investigate the inherent properties of its latent space. 
\section{Preliminaries}
\label{sec:preliminaries}

 

\subsection{Problem formulation}




Our goal is to invert a 3D mesh in the latent space of a text-to-3D flow model. Namely, our inputs are: (1)~a 3D shape $\mathcal{X}$ embedded in the representation space of the flow model as a 3D-VAE latent $z_0$, (2)~the conditioning signal $C$ paired with the source shape (which takes the form of a text prompt $\mathcal{P}_{src}$), and (3)~a trained text-to-3D flow model $v_\theta(x, t, C)$ that can generate 3D shape latents $z$ from noise.

\noindent Our output is a noisy latent $z_1$, such that when re-integrating the generation process, the generated $\hat{z_0}$ matches $z_0$ closely. This inversion process is necessary as we want the mesh to be in a representation that the generative model can manipulate. When the process of inversion is successful, editing becomes a seamless operation: we can simply use an edit prompt $\mathcal{P}_{edit}$ and $\hat{z_0}$ to obtain the desired shape.




\subsection{Rectified Flow and Inversion}
\label{subsec:flow_model}

\textbf{Rectified flow models}~\cite{lipman2022flow, liu2022flow} learn a velocity field $v_\theta(x_t, t)$ that transports a simple prior $p_1$ to a target data distribution $p_0$ by integrating the ordinary differential equation (ODE) $dx_t/dt = v_t(x_t)$. Sampling integrates the ODE from $t=1$ to $t=0$. Conversely, inversion maps a data latent $z_0$ back to noise $z_1$ by integrating the ODE forward ($t: 0 \to 1$). Following a deterministic baseline analogous to DDIM inversion~\cite{ddim}, we utilize forward Euler steps:
\begin{equation}
    z_{t_{i+1}} = z_{t_i} + (t_{i+1} - t_i) \hat{v}_\theta(z_{t_i}, t_i, C),
    \label{eq:forward_euler}
\end{equation}
where $\hat{v}_\theta$ is the velocity field modified by classifier-free guidance (CFG)~\cite{cfg}. Specifically, given a conditioning signal $C$ and guidance scale $w$, the guided velocity is computed as $\hat{v}_\theta(z_t, t, C) = v_\theta(z_t, t, \emptyset) + w(v_\theta(z_t, t, C) - v_\theta(z_t, t, \emptyset))$, where $\emptyset$ represents the null condition.\\
\\
\noindent \textbf{Inversion} is the reverse process of generation: mapping any given input $z$ to the noise space. Given a data latent $z_0$, one can invert the rectified flow to recover the corresponding noise latent $z_1$ by integrating the ODE in the forward direction using Eq.~\ref{eq:forward_euler}
with $t_0 = 0$ and $t_N = 1$ \cite{lai2025principles, rout2024rfedit}. We call this process Euler inversion.\\
\\
\noindent \textbf{Null-Text Inversion (NTI)}~\cite{nti} improves reconstruction quality when text-shape alignment is imperfect. Since direct Euler inversion can result in inaccurate reconstructions, NTI optimizes the unconditional embeddings along the forward trajectory. During sampling, these optimized null embeddings ensure the generation trajectory closely matches the inverted reference path, preserving the original shape's identity while enabling subsequent prompt-based editing.

\begin{equation} v_\theta(x_t, t) = (1 + \omega)\, v_\theta(x_t, t, C) - \omega\, v_\theta(x_t, t, \emptyset), \label{eq:cfg_equation} \end{equation}

\subsection{3D Generative Backbone}
Our study utilizes TRELLIS~\cite{xiang2024structured}, a state-of-the-art open-source framework for high-fidelity \textit{text-to-3D} generation. We do not consider recent iterations of the method, including TRELLIS.2~\cite{xiang2025trellis2}, as its improvements primarily target the image-to-3D branch, which falls outside the scope of our study. TRELLIS generates assets in two stages:

\label{subsec:backbone}
\noindent \textbf{Sparse Structure Generation.} A first rectified flow model $\mathcal{G}_S$ produces a low-dimensional latent $z_0 \in \mathbb{R}^{8 \times 16 \times 16 \times 16}$, which is decoded into a coarse $64^3$ voxel-based structure. This stage defines the global geometry of the asset. Any mesh can be projected into this space by voxelization and encoding via the pre-trained VAE (more details in Supp. Sec. B).
\\
\noindent \textbf{Structured Latent Generation.} A second model $\mathcal{G}_L$ refines the coarse structure by generating high-dimensional sparse latents (SLATs), which are necessary for creating the resulting texture details. These latents are then decoded into the final 3D representations, such as meshes, radiance fields or Gaussian splats. 
Our study primarily operates within the latent space of the structure model $\mathcal{G}_S$. Modifying $\mathcal{G}_L$ predominantly results in texture-level updates with only marginal changes to the underlying geometry (see Supp. Sec. B).


\section{Expressivity of Language and Geometry}
\label{sec:expressivity}

Our key observation is that the assumption that textual instructions can reliably navigate the latent space of 3D models often fails when prompts fall out-of-distribution. We first quantify this failure by measuring the model's responsiveness to different semantic categories and identifying where generation collapses into ``sink traps''. We then evaluate how different text prompts impact inversion stability compared to image-based models. Finally, we demonstrate that by bypassing text-conditioning, the underlying generative prior can be used to create and edit complex shapes. 

\subsection{Language Diversity}
\label{subsec:language-diversity}

\begin{figure}
    \centering
    \includegraphics[width=0.95\linewidth]{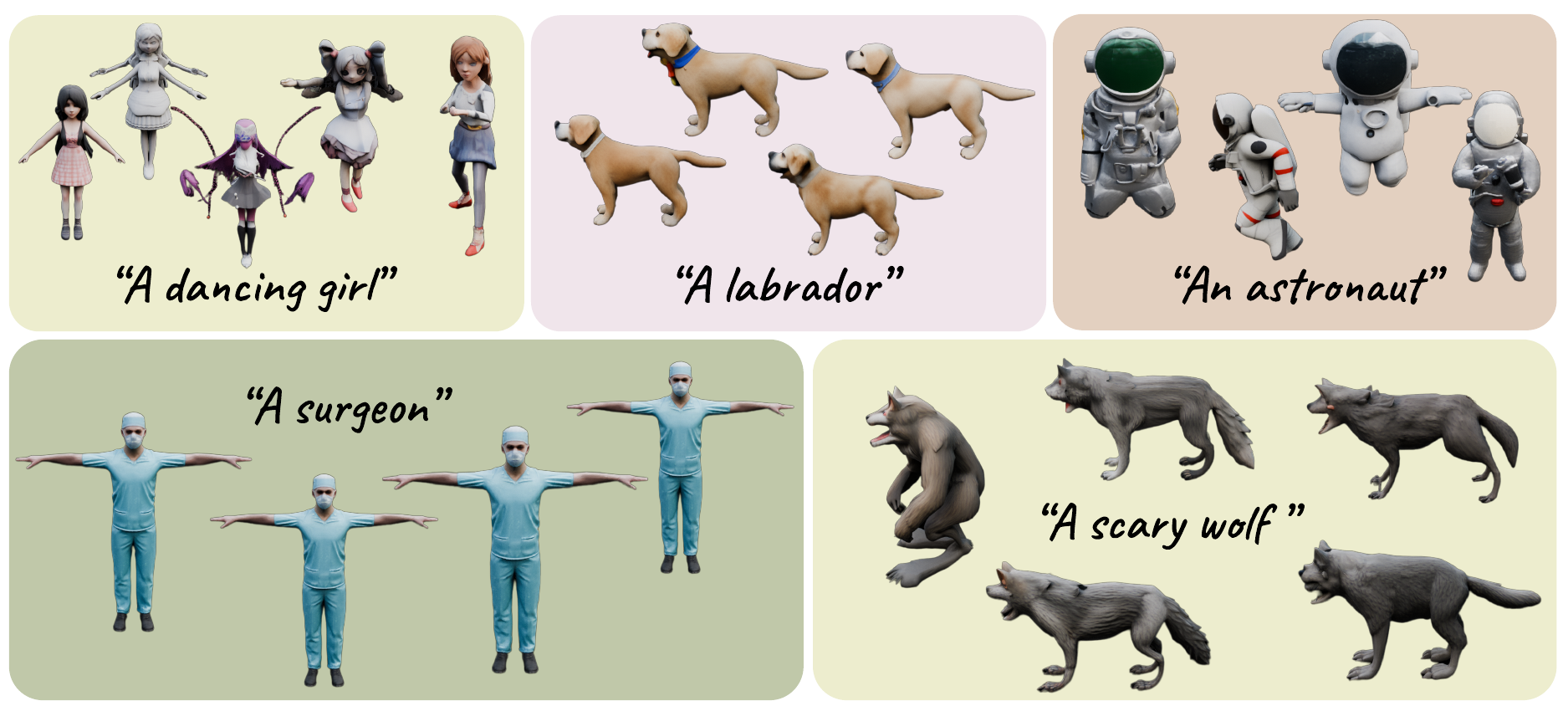}
    \caption{We generate multiple assets using TRELLIS~\cite{xiang2024structured} across diverse character classes (\eg, surgeon, astronaut) while varying specific prompt attributes (\eg, ``[Class] in a sitting pose'' vs. ``sprinting with arms swinging back and forth''). Despite these targeted variations, we observe significant mode collapse, where the model converges to a nearly identical geometry and texture for some classes, failing to modify the final result as requested by the text prompt.}
    \label{fig:sink_examples}
\end{figure}

The generative fidelity of 3D generative models is non-uniform across the prompt distribution. We observe that for specific semantic categories, the model exhibits a form of localized model collapse. In these regions, the model loses sensitivity to textual variations and converges toward a generic shape. This is qualitatively shown in Fig.~\ref{fig:sink_examples}, where diverse prompts regarding specific subjects yield near-identical topology. Note the variation across semantic categories.  

To quantify this behaviour, we perform a diversity analysis using SigLIP~\cite{tschannen2025siglip2multilingualvisionlanguage}. We generate a collection of prompts $\mathcal{T} = \{t_1, \dots, t_n\}$ centered on specific semantic anchors (\eg, astronaut, labrador) using Gemini 3 Pro~\cite{Google2025Gemini3} and create associated shapes with TRELLIS~\cite{xiang2024structured}. 

For each generated asset, we obtain a set of multi-view renders $\mathcal{I}_i$ covering a $360^\circ$ orbit. Let $\Phi_{\text{vis}}(\cdot)$ and $\Phi_{\text{txt}}(\cdot)$ denote the SigLIP image and text encoders, respectively. We define the average pairwise distance for both modalities as:
\begin{equation}
\begin{aligned}
\Delta_{\text{vis}} &= \frac{1}{N(N-1)} \sum_{i \neq j} \Big(1 - \cos\big(\Phi_{\text{vis}}(\mathcal{I}_i), \Phi_{\text{vis}}(\mathcal{I}_j)\big)\Big) \\
\Delta_{\text{txt}} &= \frac{1}{N(N-1)} \sum_{i \neq j} \Big(1 - \cos\big(\Phi_{\text{txt}}(t_i), \Phi_{\text{txt}}(t_j)\big)\Big)
\end{aligned}
\end{equation}

The diversity ratio $R = \Delta_{\text{vis}} / \Delta_{\text{txt}}$ characterizes the model's responsiveness to linguistic shifts. This ratio is closely related to the Lipschitz constant of the generative model with respect to text inputs, where the main difference is that the cosine distance does not necessarily define a metric space. 

\begin{wraptable}[13]{r}{0.55\linewidth}    \centering
    \vspace{-2.8em} 
    \caption{{Values of $\Delta_{\text{vis}}$, $\Delta_{\text{txt}}$, and $R$ across different subjects for TRELLIS~\cite{xiang2024structured}, LN3Diff~\cite{ln3diff} (LN), and GaussianAnything~\cite{gaussiananything} (GA). Most subjects exhibit $R < 1$.}}
    \scriptsize
    \begin{tabular*}{\linewidth}{@{\extracolsep{\fill}} lccccc }
        \toprule
        & \multicolumn{3}{c}{TRELLIS} & {LN} & {GA} \\
        \cmidrule(lr){2-4} \cmidrule(lr){5-5} \cmidrule(lr){6-6}
        Subject & $\Delta_{\text{vis}}$ & $\Delta_{\text{txt}}$ & $R$ & {$R$} & {$R$} \\
        \midrule
        Cute puppy   & 0.376 & 0.394 & 0.990 & {0.970} & {0.538} \\
        Surgeon      & 0.196 & 0.377 & 0.515 & {0.806} & {0.613} \\
        Husky        & 0.381 & 0.427 & 0.799 & {0.651} & {0.496} \\
        Dancing girl & 0.512 & 0.404 & 1.205 & {0.626} & {0.569} \\
        Car          & 0.420 & 0.367 & 1.124 & {0.417} & {0.452} \\
        Scary wolf   & 0.156 & 0.270 & 0.640 & {1.041} & {0.707} \\
        Astronaut    & 0.260 & 0.461 & 0.638 & {0.861} & {0.529} \\
        Labrador     & 0.441 & 0.489 & 0.788 & {0.697} & {0.389} \\
        \bottomrule
    \end{tabular*}
    \label{tab:siglip_diversity}
\end{wraptable}

A low ratio ($R \ll 1$) indicates that the model is operating in a sparse region of the joint embedding space, where noticeable changes in the text prompt lead to small (or no) visual changes. This behavior can be attributed to the text capability of the generative model being effectively out-of-distribution (OOD) relative to the learned geometric priors. In such cases, the conditional signal fails to steer the generation, causing the output to collapse toward a generic canonical shape. Our quantitative results in Tab.~\ref{tab:siglip_diversity} suggest that the text-to-shape manifold is non-uniformly populated. Some specific semantic groups reside in densely mapped regions that facilitate high diversity, while others encounter sink regions with little to zero diversity. {To confirm this phenomenon generalizes beyond TRELLIS, we additionally evaluated LN3Diff~\cite{ln3diff} and GaussianAnything~\cite{gaussiananything} models. As shown in Tab.~\ref{tab:siglip_diversity}, both models consistently exhibit $R < 1$ across the majority of categories. We also evaluated models prior to 2024\cite{zhang2024gaussiancubestructuredexplicitradiance, jun2023shapegeneratingconditional3d, ren2024xcubelargescale3dgenerative}, however their quantitative performance was too poor to generate plausible results for many input prompts.}

This poses a significant challenge for 3D shape editing: in general, the user does not have access to the ``ground truth prompt'' of the shape, but rather an approximate prompt based on their observation, or the help of an auxiliary tool (\eg, LLM). Because of the existence of sparse regions in the joint embedding space, the approximate prompt will likely be out-of-distribution (OOD). {Therefore, using an approximate prompt will unlikely yield the desired result as the model will become unresponsive to it.}

\begin{figure}[!htbp]
    \centering
    \includegraphics[width=0.95\linewidth]{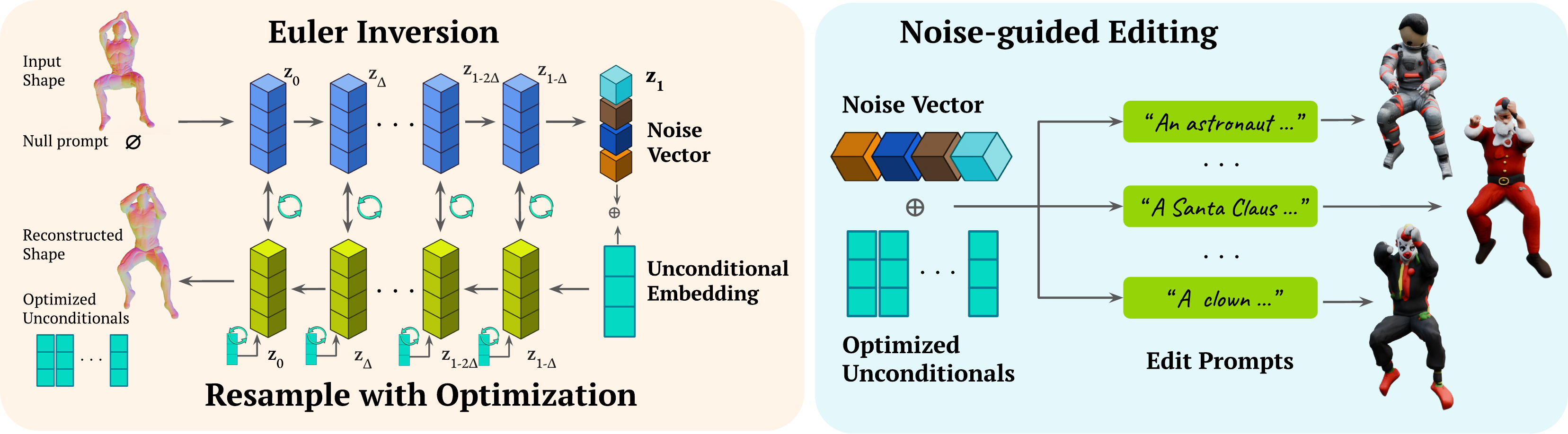}
    \caption{Our unconditional 3D shape inversion (Left) and text-driven editing (Right) pipelines. We invert an arbitrary input shape by using an empty prompt and refining its embedding via NTI optimization~\cite{nti}. Remarkably, this unconditional inversion strategy not only yields superior reconstructions, but the resulting noisy latent and optimized embedding also supports powerful, open-vocabulary editing of the input 3D shape.}
    \label{fig:editing_pipeline}
\end{figure}

\subsection{Out-of-Distribution Prompts}
\label{subsec:ood-prompts}

As inversion requires a source prompt that corresponds to the initial geometry, when the user provides an OOD prompt relative to the shape, the joint embedding fails to align, and the resulting reconstruction will fail. Worse, any downstream task, such as editing, will be affected, and the output will not correspond to the user's intent. This behaviour differs from T2I models, where it has been noted that providing ``any reasonable and editable prompt'' is sufficient for robust inversion and editing~\cite{nti}. 

\begin{figure}[!htbp]
\centering
\vspace{-20pt}
\begin{minipage}[c]{0.48\columnwidth}
\vspace{20pt}
\centering
\includegraphics[width=\linewidth]{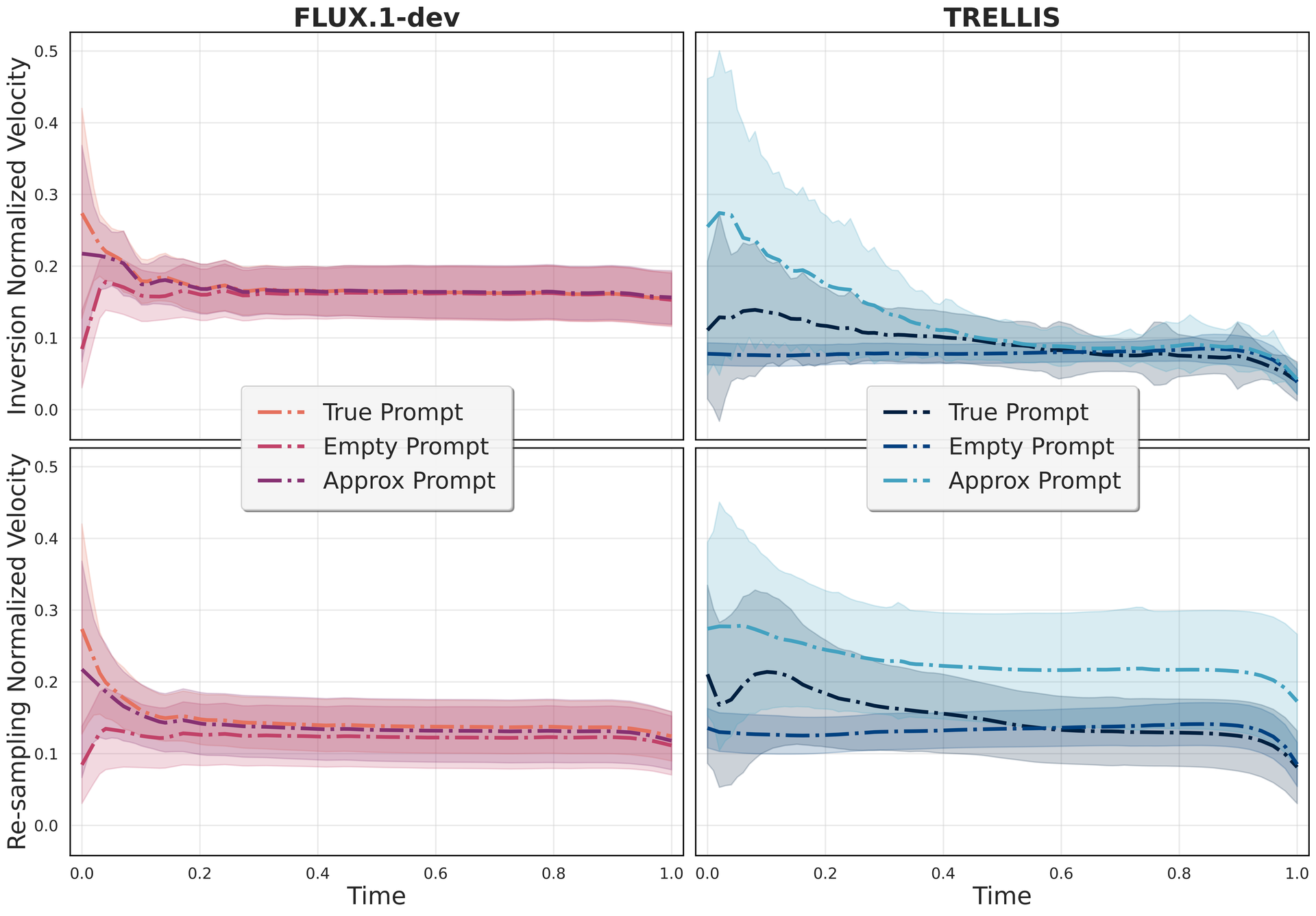}
\vspace{-15pt}
\captionof{figure}{The velocity norm of FLUX.1~\cite{labs2025flux1} remains stable across different prompt types, whereas TRELLIS~\cite{xiang2024structured} exhibits large variations.}
\label{fig:flux_vs_trellis}
\end{minipage}
\hfill
\begin{minipage}[c]{0.48\columnwidth}
    \vspace{0pt}
    \centering
    \captionof{table}{Effect of prompt types on inversion quality across different models.}
    \label{tab:prompt_type_influence}
    \vspace{5pt}
    \resizebox{\linewidth}{!}{%
    \begin{tabular}{lccc}
    \toprule
     & True Prompt & Approximate Prompt & Empty Prompt \\
    \midrule
    \multicolumn{4}{c}{\textbf{Stable Diffusion v1.4}~\cite{rombach2022sd}} \\
    PSNR $\uparrow$  & 15.00 $\pm$  2.43 & 14.61 $\pm$ 3.45 & \textbf{15.66 $\pm$  2.57} \\
    LPIPS $\downarrow$ & \textbf{0.56 $\pm$  0.07} & 0.59 $\pm$  0.08 & 0.57 $\pm$  0.07 \\
    \midrule
    \multicolumn{4}{c}{\textbf{FLUX.1 dev}~\cite{labs2025flux1}} \\
    PSNR $\uparrow$ & 10.32 $\pm$ 2.55 & 10.62 $\pm$  6.12 & \textbf{15.12 $\pm$  3.33} \\
    LPIPS $\downarrow$ & 0.57 $\pm$ 0.169 & 0.58 $\pm$ 0.19 & \textbf{0.48 $\pm$  0.13} \\
    \midrule
    \multicolumn{4}{c}{\textbf{TRELLIS}~\cite{xiang2024structured}} \\
    L1 $\downarrow$ & 17.75 $\pm$  34.59 & 76.55 $\pm$  73.35 & \textbf{5.40 $\pm$  14.15} \\
    \midrule
    \multicolumn{4}{c}{\textbf{LN3Diff}~\cite{ln3diff}} \\
    L1 $\downarrow$ & 1.75 $\pm$ 0.98 & 1.82 $\pm$ 1.16 & \textbf{0.05 $\pm$ 0.07} \\
    \midrule
    \multicolumn{4}{c}{\textbf{GaussianAnything}~\cite{gaussiananything}} \\
    L1 $\downarrow$ & 0.32 $\pm$ 0.45 & 0.99 $\pm$ 0.47 & \textbf{0.19 $\pm$ 0.06} \\
    \bottomrule
    \end{tabular}%
    }
\end{minipage}
\end{figure}

To validate this, we conducted an inversion and resampling experiment on TRELLIS-generated assets using three prompt conditions: the true prompt, a two-word approximate variant, and an empty prompt. Reconstruction accuracy was evaluated using the $L_1$ score on the coarse voxel structure produced by the geometric stage $\mathcal{G}_S$. As shown in Tab.~\ref{tab:prompt_type_influence}, approximate prompts led to poor reconstructions in TRELLIS, whereas in comparison trials with Stable Diffusion~\cite{rombach2022sd} and FLUX.1~\cite{labs2025flux1}, reconstructions with approximate prompts remained comparable to those using original prompts.

We further analyze this instability by computing the norm of the predicted velocity during inversion. In T2I models, the sequence of score norms across noise levels can detect OOD data samples~\cite{ooddetect}. For the flow-based formulation of TRELLIS, we evaluate the equivalent sequence of velocity norms $\|\mathbf{v}_t\|$ across timesteps~\cite{lai2025principles}. Our results show that approximate prompts result in substantially higher velocity norms, signaling an attempt by the model to escape low-density regions. 

This analysis highlights both the difficulty of 3D-based inversion and the lack of previous purely text-driven editing methods. Indeed, as we show, because they become OOD prompts, using approximate prompts as advocated in image-based approaches~\cite{nti}, is not suited for text-to-3D models, even state-of-the-art ones like TRELLIS.


Instead, we find that the \textit{empty prompt} instead promotes substantially smaller velocity norms and more stable trajectories (Fig.~\ref{fig:flux_vs_trellis} and Table~\ref{tab:prompt_type_influence}). {In classifier-free guidance, the $w$-difference term between conditional and unconditional velocities can act as an unstable, high-variance gradient when the conditioning is highly out-of-distribution (\eg, from an approximate prompt). Using an empty prompt eliminates this OOD guidance, keeping the trajectory stable.} This suggests that combining the null prompt with a shape is a more reliable strategy for embedding such pairs into the model's latent space, {and also enables stable text-based editing when combined with inversion.} 

\subsection{Geometric Expressivity and Text-Editing}
\label{subsec:geo-expressivity}

\begin{figure}[t!]
    \centering
    \begin{minipage}{0.85\columnwidth} 
        \centering
        \captionsetup[subfigure]{font=scriptsize}
        \begin{subfigure}[b]{0.18\textwidth}
            \centering
            \includegraphics[width=\textwidth]{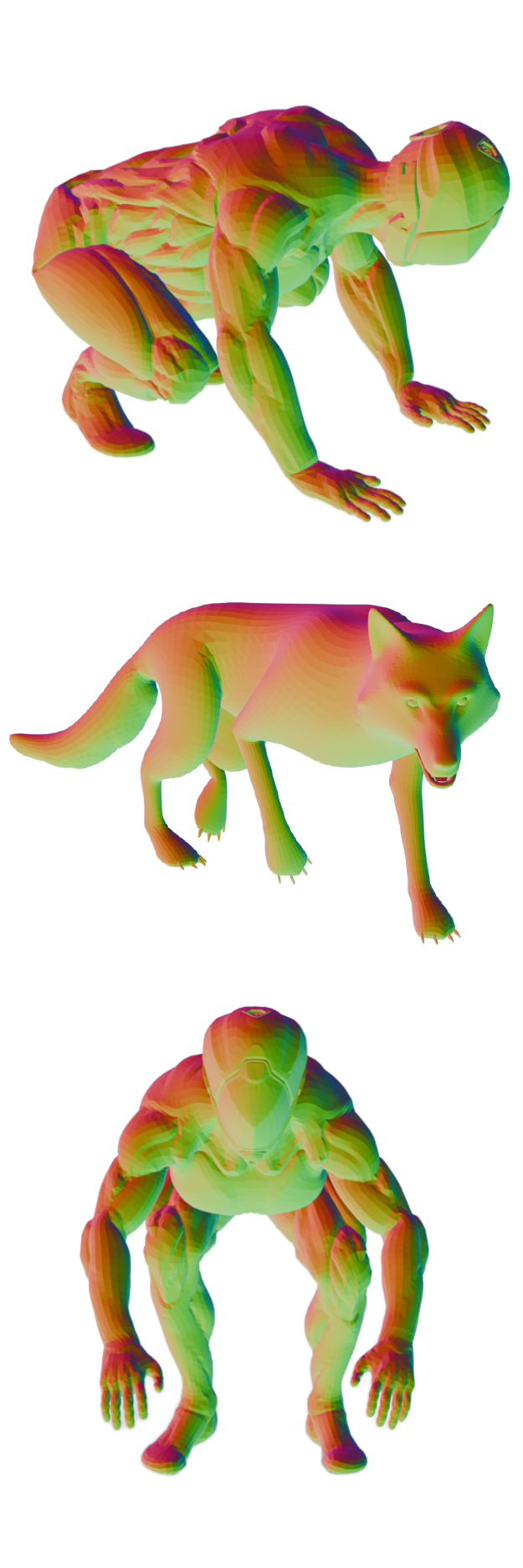}
            \caption{Target}
            \label{subfig:gt_mesh_prompt_impact}
        \end{subfigure}%
        \hfill
        \begin{subfigure}[b]{0.18\textwidth}
            \centering
            \includegraphics[width=\textwidth]{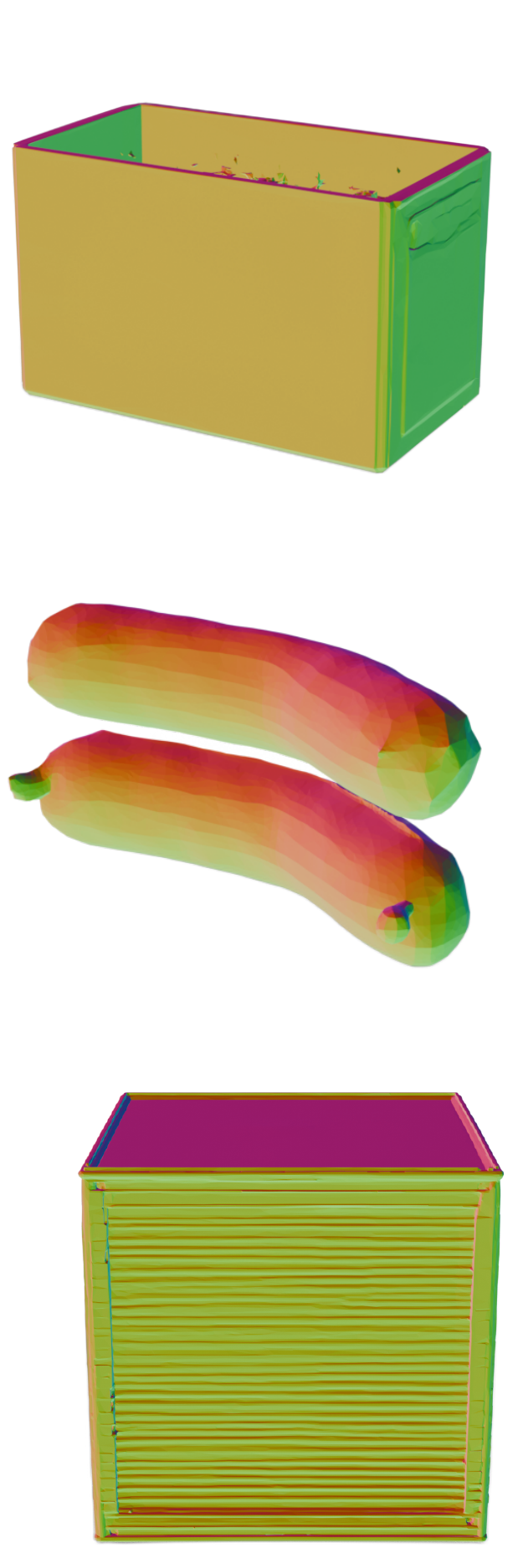}
            \caption{Euler + $\tilde{\mathcal{C}}$}
            \label{subfig:eulerp}
        \end{subfigure}%
        \hfill
        \begin{subfigure}[b]{0.18\textwidth}
            \centering
            \includegraphics[width=\textwidth]{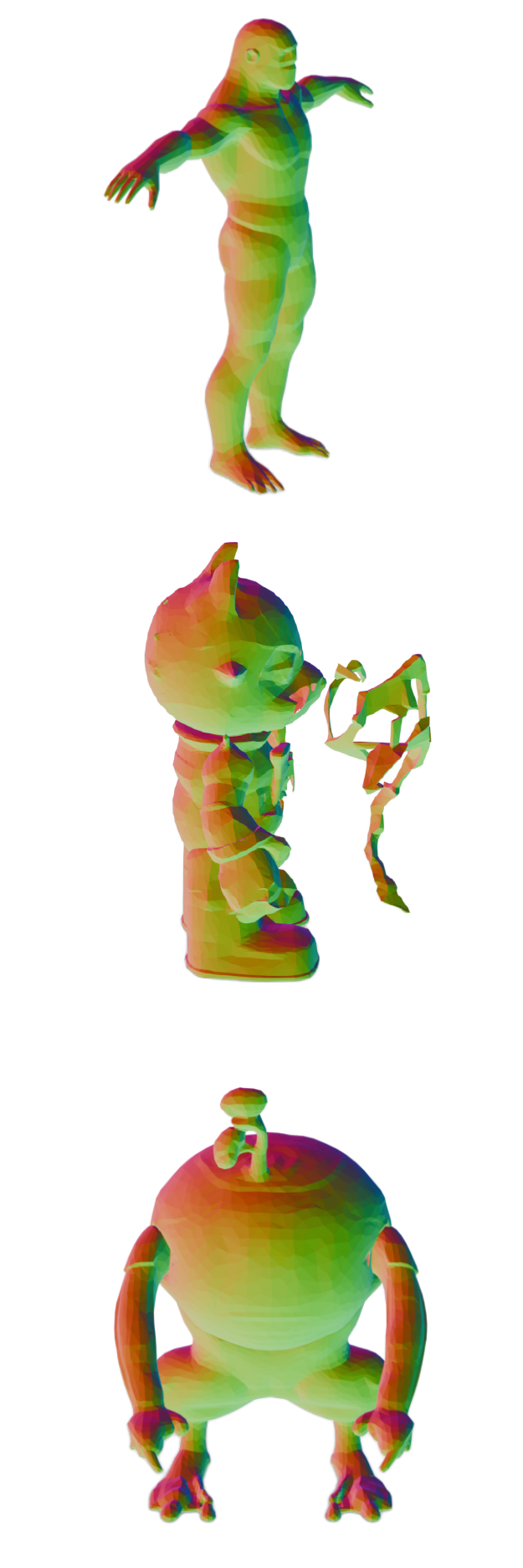}
            \caption{NTI \cite{nti}}
            \label{subfig:nullp}
        \end{subfigure}%
        \hfill
        \begin{subfigure}[b]{0.18\textwidth}
            \centering
            \includegraphics[width=\textwidth]{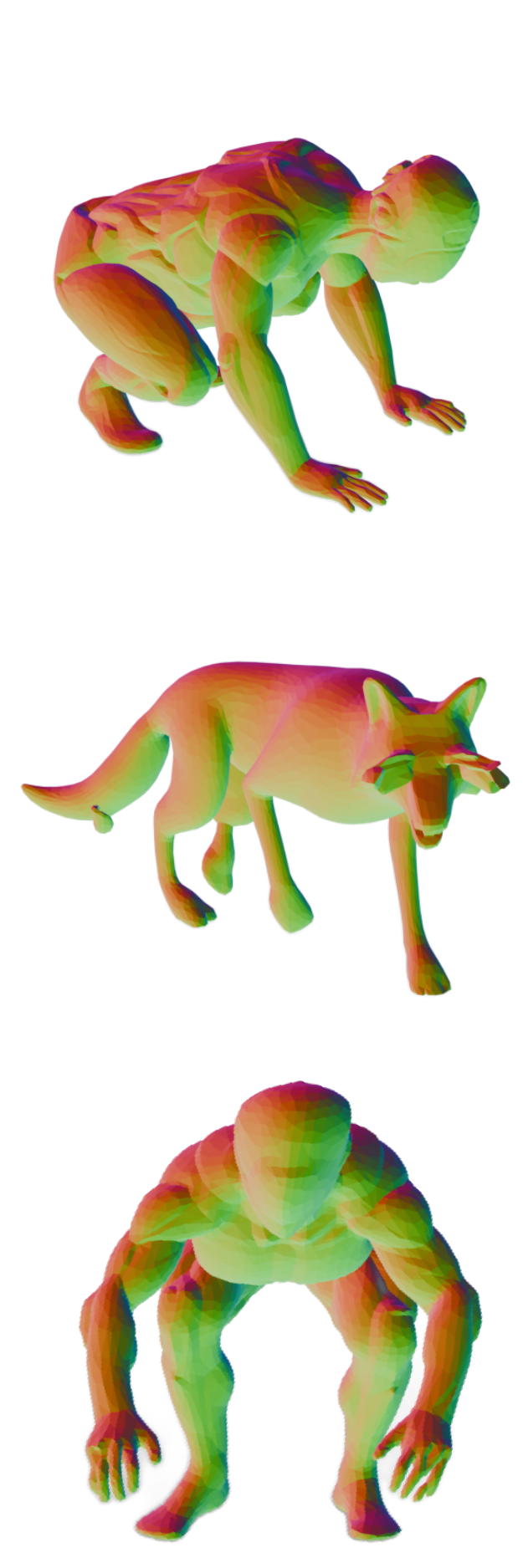}
            \caption{Euler + $\emptyset$}
            \label{subfig:euler_empty_prompt}
        \end{subfigure}%
        \hfill
        \begin{subfigure}[b]{0.18\textwidth}
            \centering
            \includegraphics[width=\textwidth]{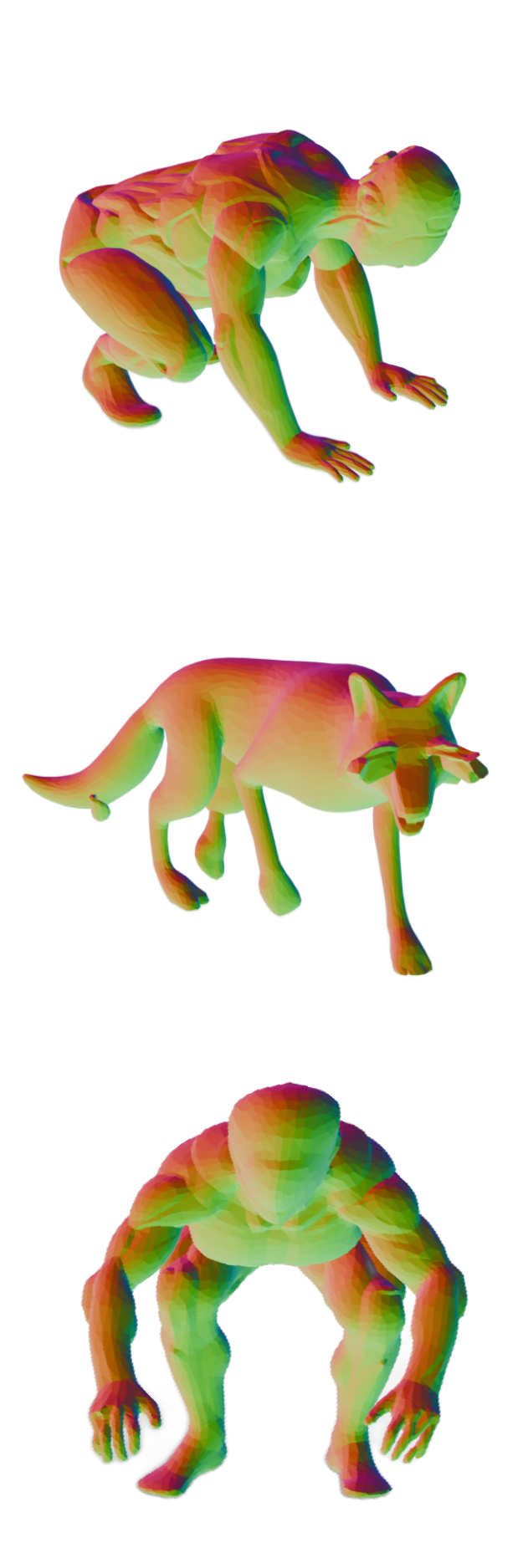}
            \caption{NTI + $\emptyset$}
            \label{subfig:null_empty_prompt}
        \end{subfigure}
    \end{minipage}
    \caption{Comparison of shape reconstruction across inversion methods. (a) Target shape. (b) Euler inversion with approximate text prompt. (c) NTI inversion with approximate text prompt. (d) Euler inversion with empty prompt $\emptyset$. (e) NTI inversion with empty prompt $\emptyset$. When using approximate text prompts (b-c), both methods fail to accurately reconstruct the target shape. In contrast, inverting with an empty prompt (d-e) enables both methods to achieve high-fidelity reconstruction.}
    \label{fig:recon_accuracy_and_approx_prompts}
\end{figure}

Building on the finding that null prompts coupled with shape embeddings mitigate OOD drift, we utilize this stability to probe the model’s inherent geometric expressivity. As illustrated in  Fig.~\ref{fig:recon_accuracy_and_approx_prompts} and in the experiment section~\ref{sec:experiments}, using an approximate prompt $\tilde{\mathcal{C}}$ often leads to implausible geometries or artifacts, likely due to the latent trajectory drifting into sparse density regions during inversion.  This capability demonstrates that unconditional embeddings can guide the model to recreate complex shapes where finding the correct text-conditioning would be very challenging. We interpret the phenomenon as a clear ``expressivity mismatch'': the model's geometric capacity far exceeds its language-constrained expressivity, proving that the latent space contains a wealth of geometric detail accessible only through unconditional guidance. 

\begin{figure}[!htbp]
  \centering
   \includegraphics[width=0.95\linewidth]{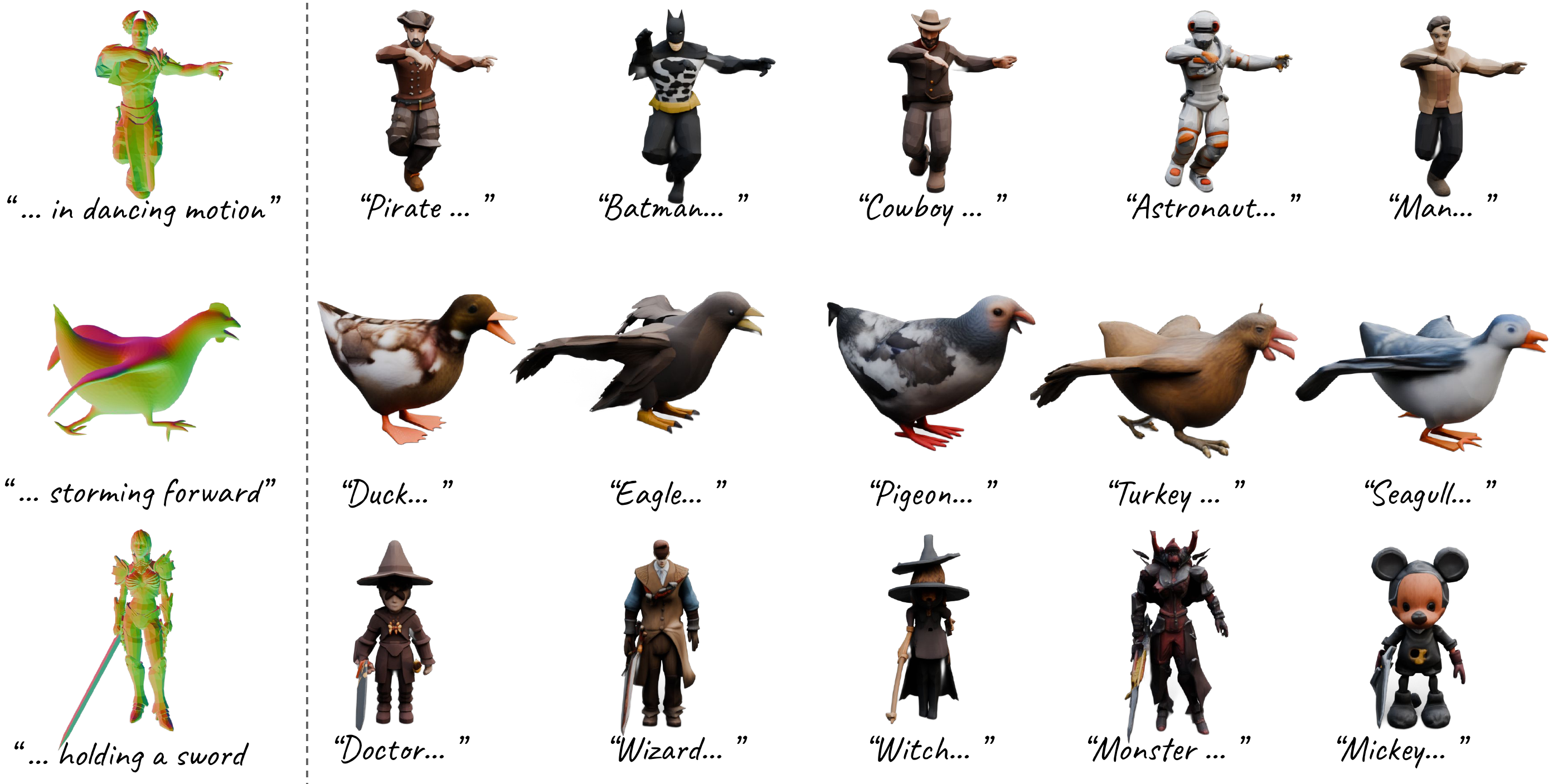}
   \caption{Open-vocabulary edits (right) on non-rigid 3D shapes (left) inverted by our method within the unconditional latent space of the TRELLIS 3D generative model. Our inversion method reliably reconstructs arbitrary shapes and enables semantic edits.}
   \label{fig:our_editing}
   
\end{figure}

\noindent \textbf{Text-guided editing of arbitrary shapes.} Our text-driven editing pipeline leverages the unconditional distribution (see Fig.~\ref{fig:editing_pipeline} for an overview). Because we invert arbitrary shapes, our approach enables text-based retargeting. Namely, given a shape $\mathcal{X}$, the objective is to re-target the source pose to diverse target characters (see Fig. \ref{fig:our_editing}). For example, if the source shape is a demon figure dancing, we can use edit prompts such as $\mathcal{P}_{edit}$ ``princess dancing'' or ``cowboy dancing''. We first invert $\mathcal{X}$ into the TRELLIS latent space using a null prompt ($\mathcal{P}_{src} = \emptyset$). To perform an edit, we resample from the inverted noise latent using the edit prompt ($\mathcal{P}_{edit}$). We optionally apply null-text optimization~\cite{nti} to optimize the unconditional embeddings along the sampling trajectory, which allows the edit to maintain further structural integrity to the original shape. 

\section{Experiments}
\label{sec:experiments}

\subsection{Experimental setup}

\noindent \textbf{Implementation} For inversion, we use classifier-free guidance (CFG) scale of 5 and 50 sampling timesteps. We perform inversion exclusively on the first-stage flow model $\mathcal{G}_S$, which dictates the global geometric structure, while keeping the second-stage model $\mathcal{G}_L$ fixed and seeded to isolate the source of variation. When employing Null-Text Inversion (NTI)~\cite{nti}, we use the Adam optimizer \cite{kingma2017adam} with a learning rate of $1\times10^{-4}$ and 10 inner optimization steps per sampling step, conducted on a single L40S GPU. We measure reconstruction quality using an L1 metric on the decoded latents, defined as $L1 = |\hat{\mathbf{x}}_0 - \mathcal{D}(\mathcal{E}(\mathbf{x}_0))|$, where $\mathcal{E}$ and $\mathcal{D}$ represent the VAE encoder and decoder. This ensures we account for the inherent VAE reconstruction bias. To capture finer surface details and perceptual fidelity, we also compute LPIPS~\cite{zhang2018unreasonable} scores between rendered normals of the original and reconstructed meshes across 10 distinct viewpoints.\\
\\
\noindent \textbf{Datasets} We evaluate our method using two datasets: 200 non-rigid humanoid and animal characters from the DT4D dataset \cite{li2021dt4d} to assess open-vocabulary semantic edits on complex, out-of-distribution (OOD) shapes, and a subset of 80 TRELLIS-generated shapes. The latter is used strictly for comparisons to other methods, as existing methods like VoxHammer \cite{li2025voxhammer} suffer from latent value explosion on DT4D (see Supp. Sec. D for precise dataset preparation). The detailed creation of the TRELLIS-generated shapes, as well as the setups of Sections \ref{subsec:language-diversity} and \ref{subsec:ood-prompts} are detailed in the supplementary material. 

\subsection{Evaluation of geometric expressivity}

To evaluate the geometric expressivity of TRELLIS, we compare four configurations: Euler inversion and NTI, each paired with either an approximate prompt $\tilde{\mathcal{C}}$ (e.g., ``A horse galloping'') or an empty prompt $\emptyset$. Results in Supp. Mat. (Table 1) demonstrate that using an approximate prompt can be detrimental to reconstruction quality; while NTI with $\tilde{\mathcal{C}}$ improves reconstruction over Euler with $\tilde{\mathcal{C}}$, it remains significantly worse than using an empty prompt. Notably, when using $\emptyset$, both Euler and NTI achieve similar, high-accuracy results.

\begin{figure}[t]
  \centering
   \includegraphics[width=0.85\linewidth]{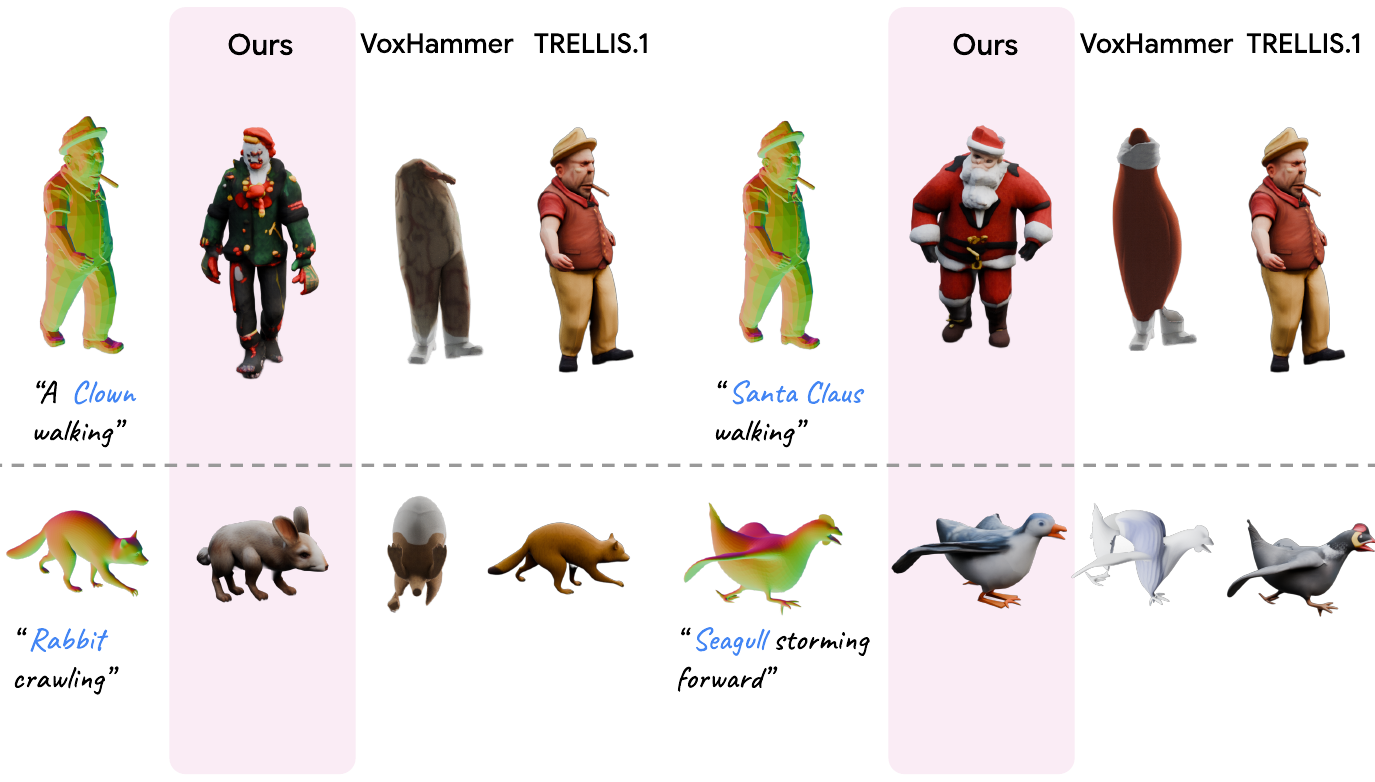}
   \caption{Comparison of our editing with native 3D space baselines: the state-of-the-art VoxHammer, and the TRELLIS second stage edit. VoxHammer fails to create plausible shapes due to the inversion failure, while TRELLIS proposes only texture changes. Ours is the only method to provide meaningful retargeting based on the editing prompts.}
   \label{fig:compare_trellis_voxhammer}
\end{figure}

As illustrated in Fig.~\ref{fig:recon_accuracy_and_approx_prompts}, using an approximate prompt $\tilde{\mathcal{C}}$ often leads to implausible geometries or artifacts, likely due to the latent trajectory drifting into sparse density regions during inversion. In contrast, Euler inversion with a null prompt consistently achieves the highest reconstruction accuracy, successfully capturing the complex, non-rigid topologies inherent in DT4D. This capability demonstrates that unconditional embeddings can guide the model to recreate complex shapes where finding in-distribution text-conditioning would be very challenging. 

\subsection{Text-Guided Editing Evaluation}

We now evaluate editing capabilities by creating edit prompts $\mathcal{P}_{edit}$  that modify appearance while preserving the original pose and structure. 
\paragraph{\textbf{Baselines.}}
We compare our approach against the native editing procedure of TRELLIS \cite{xiang2024structured}, the text-based pipeline of VoxHammer \cite{li2025voxhammer}, and two image-prior baselines: Vox-E \cite{sella2023voxetextguidedvoxelediting} and Instant3DiT \cite{barda2024instant3ditmultiviewinpaintingfast} (visual comparisons in Supp. Fig. 7. We exclude full-scene editing methods to ensure a fair, direct quantitative comparison with our mesh-based object editing. Additionally, TRELLIS' native editing retains coarse structures by re-running only its second-stage model ($\mathcal{G}_L$); however, it primarily modifies texture with negligible geometric updates (Fig.~\ref{fig:compare_trellis_voxhammer}). Furthermore, VoxHammer fails on 67\% of DT4D dataset samples due to inversion drift causing latent value explosion (rendering shapes undecodable by the VAE). Even when successful, VoxHammer requires manual 3D masking, frequently produces implausible geometry, and struggles to modify subjects while preserving their original pose.

\paragraph{\textbf{Quantitative and Visual Results.}} 
To evaluate editing quality, we sample 30 rendered views per edited sample. Table~\ref{tab:siglip_scores} reports alignment with the target prompt $\mathcal{P}_{edit}$ using SigLIP \cite{tschannen2025siglip2multilingualvisionlanguage}, CLIP scores, and LLaVA-1.5-7B \cite{liu2023visualinstructiontuning} for VLM-based evaluation (1--5 scale). Starting from inverted noise (Euler + $\emptyset$ and NTI + $\emptyset$), our method consistently outperforms all baselines. Beyond visual quality, our approach offers a substantial computational advantage. Both Instant3DiT~\cite{barda2024instant3ditmultiviewinpaintingfast} and VoxHammer~\cite{li2025voxhammer} require manual mask creation, with VoxHammer additionally rendering over 150 views to extract multi-view DINOv2 \cite{oquab2024dinov2learningrobustvisual} features. Meanwhile, Vox-E~\cite{sella2023voxetextguidedvoxelediting} relies on notoriously slow per-scene optimization, taking over an hour per edit. In contrast, our pipeline requires only input mesh voxelization and three generative sampling passes (averaging 9s per edit). This drastically improves computational efficiency while maintaining higher structural integrity.

\subsection{Geometric Consistency}
\begin{figure}[t]
  \centering
   \includegraphics[width=0.9\linewidth]{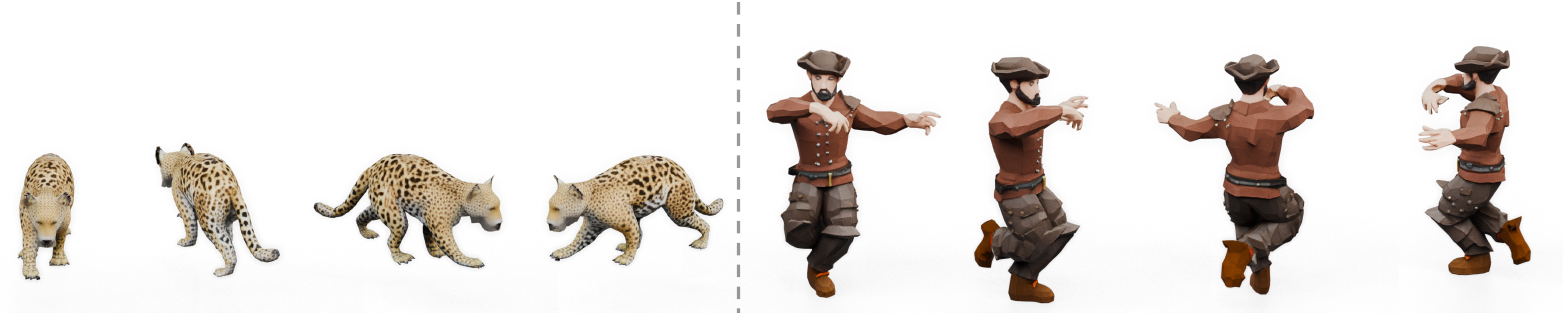}
   \caption{Multiview consistency of 3D edits. Because our method operates natively in the 3D space, the provided edits are naturally view-consistent.}
   \label{fig:mv_images}
\end{figure}

Despite enabling global geometric modifications, our edits remain strictly 3D-consistent. We provide multiview renders of our results in Fig.~\ref{fig:mv_images}. These renders confirm that our method maintains structural coherence and rendering stability across all viewing angles. Further videos are provided in the supplementary material.


\section{Limitations and future work}

\begin{figure}[ht]
  \centering
  \begin{minipage}[c]{0.55\linewidth}
    \centering
    \includegraphics[width=\linewidth]{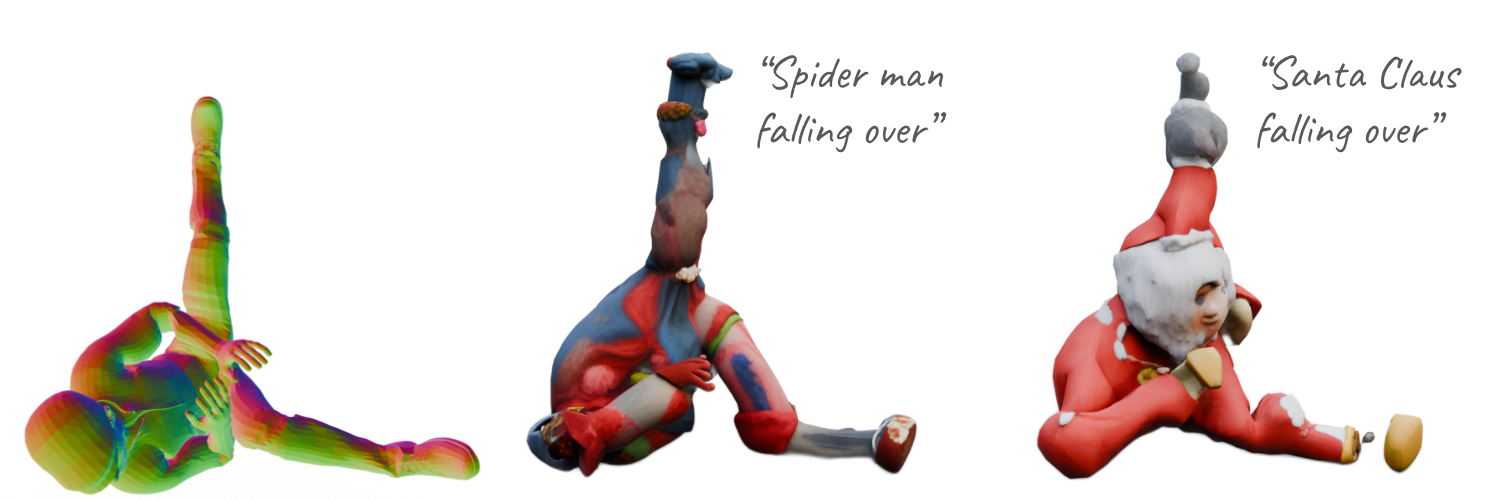}
    \caption{Bottom-left: input shape in a dancing pose on the floor. Bottom-right: two edits. The edits showcase failure modes, likely because the pose is geometrically out of distribution.}
    \label{fig:merged_edits}
  \end{minipage}
  \hfill 
  \begin{minipage}[c]{0.40\linewidth}
    \centering
    \captionof{table}{Quantitative evaluation of the edited assets. $^\dagger$ indicates methods requiring image-based priors.}
    \label{tab:siglip_scores}
    \resizebox{\linewidth}{!}{%
    \begin{tabular}{lccc}
    \toprule
    Method & SigLIP $\uparrow$ & CLIP $\uparrow$ & VLM $\uparrow$ \\
    \midrule
    TRELLIS.1~\cite{xiang2024structured} & 0.0797 & 0.2489 & 2.65 \\
    VoxHammer~\cite{li2025voxhammer} & 0.0240 & 0.2207 & 1.33 \\
    Vox-E$^\dagger$~\cite{sella2023voxetextguidedvoxelediting} & 0.0250 & 0.2430 & 1.84 \\
    Instant3DiT$^\dagger$~\cite{barda2024instant3ditmultiviewinpaintingfast} & 0.0405 & 0.2434 & 1.99 \\
    \textbf{Ours} & \textbf{0.1469} & \textbf{0.2829} & \textbf{3.98} \\ 
    \bottomrule
    \end{tabular}%
    }
\end{minipage}
\end{figure}
\noindent \textbf{Limitations.}
Although our method produces high-quality edits, it is fundamentally bounded by the generative model's underlying distribution. Unconditional inversion successfully recovers the structure of highly out-of-distribution shapes, but the subsequent editing trajectories can become unstable under simple prompts. In these extreme cases, applying edit prompts with the retrieved noise latent and unconditional embeddings yields implausible geometries (Fig.~\ref{fig:merged_edits}). Because our optimization does not explicitly enforce geometric plausibility, large semantic shifts on rare poses may produce unrealistic results. \\
\\
\noindent \textbf{Future work.} First, future research can focus on improving editing robustness, notably by automatically detecting out-of-distribution and low-diversity states. Second, inversion could be applied to new downstream tasks; since it captures rich geometric cues, adapting image morphing techniques~\cite{yang2024impus} to 3D shapes would benefit shape analysis applications.

\section{Conclusion}
\label{sec:conclusion}
We present the first systematic investigation of inversion techniques for native 3D generative models, revealing fundamental differences from established image-domain methods. Our key finding is that weak text-shape alignment in 3D generative models causes standard prompt-guided inversion to drift into sparse regions of the learned distribution, resulting in poor reconstructions and implausible geometries.
To address this challenge, we propose a simple yet effective solution: leveraging the unconditional generative prior during inversion. By performing Euler inversion with an empty prompt and optimizing only the unconditional embeddings via null-text optimization, we achieve high-fidelity reconstructions of complex, out-of-distribution shapes. Remarkably, this unconditional inversion strategy not only improves reconstruction quality but also enables meaningful text-driven editing while preserving the source shape's global structure and pose. Finally, even with the rapid growth of 3D data repositories, text-to-3D generation is likely to remain data-constrained, meaning user-provided meshes may lack corresponding generative prompts. Our approach sheds light on the possibility of inversion and editing in such data-constrained regimes.

\section*{Acknowledgements}
Parts of this work were supported by the ERC Consolidator Grant ``VEGA’' (No. 101087347), as well as gifts from Ansys Inc., and Adobe Research. This project was provided with computing AI and storage resources by GENCI at IDRIS thanks to the grant 2026-AD010613760R3 on the supercomputer Jean Zay's H100 partition.


\bibliographystyle{splncs04}
\bibliography{main}

\clearpage
\appendix
\thispagestyle{plain}
\begin{center}
{\LARGE \bfseries Supplementary Material\\[0.5em]\par}
\end{center}

\vspace{1em}

In this supplementary material, we provide additional discussions and results to complement our main paper. Specifically, in Sec.~\ref{sec:implementation}, we detail the specific algorithms and parameters used for our Euler inversion and null-text optimization (Alg.~\ref{alg:nullinv}). In Sec.~\ref{sec:supp_inversion_image}, we expand our analysis of prompt sensitivity, providing qualitative and quantitative comparisons using image generative models such as FLUX and Stable Diffusion to illustrate why 3D inversion is uniquely susceptible to out-of-distribution (OOD) drift. Sec.~\ref{sec:supp_comp_trellis} compares our approach against the native TRELLIS editing procedure, demonstrating our method’s superior ability to control global geometry rather than only surface texture. We then provide details regarding our data preparation and prompt construction for the DT4D and TRELLIS datasets in Sec. ~\ref{sec:supp_data_preparation}. Finally, we provide further qualitative examples of our open-vocabulary shape editing capabilities in Sec.~\ref{sec:supp_editing_additional}.

\section{Implementation Details: Inversion and Null Text Optimization}
\label{sec:implementation}
We build on the publicly released TRELLIS~\cite{xiang2024structured} codebase and use its text-to-3D diffusion model for all experiments. 
Given an input mesh, we first convert it to the voxel representation required by the model; this follows the original data preprocessing pipeline. Note that the entire editing pipeline operates only in the 3D space, hence only the original mesh is required as input.   

To recover the noise latent associated with a given shape, we perform Euler inversion over 50 steps, integrating from $t = 0$ to $t = 1$. The conditional embedding is fixed to the empty prompt throughout inversion. 
At each step, we record the intermediate latent $z_{t_i}$, which yields both the final predicted noise latent and the entire inversion trajectory $\{z_{t_i}^{\mathrm{ref}}\}$.

Null-text optimization is used in the sampling stage and optimizes only the unconditional embeddings
used at each timestep $t_i$ from $t = 1$ to $t = 0$. The pseudo-algorithm for our NTI optimization is given in Alg.~\ref{alg:nullinv}. We initialize sampling from
$z^{\mathrm{ref}}_{t_0}$, where $t_0 = 1$. For each timestep $t_i$, we introduce an 
optimizable unconditional embedding $e^{(i)}_{\mathrm{uncond}}$. Given the current latent $z_{t_i}$, 
we compute the conditional velocity 
$v_{\mathrm{cond}} = v_\theta(z_{t_i}, t_i, \emptyset)$
and the unconditional velocity
$v_{\mathrm{uncond}} = v_\theta(z_{t_i}, t_i, e^{(i)}_{\mathrm{uncond}})$.
Classifier-free guidance is then applied to form
\[
v_{\mathrm{pred}} 
= v_{\mathrm{cond}} 
+ \omega ( v_{\mathrm{cond}} - v_{\mathrm{uncond}} ),
\]
from which the predicted next latent is obtained via a backward Euler step
\[
z^{\mathrm{pred}}_{t_{i+1}}
= z_{t_i} - (t_i - t_{i+1}) \, v_{\mathrm{pred}}.
\]
The unconditional embedding $e^{(i)}_{\mathrm{uncond}}$ is optimized using Adam for 10 inner 
iterations by minimizing the deviation between $z^{\mathrm{pred}}_{t_{i+1}}$ and the reference 
latent $z^{\mathrm{ref}}_{t_{i+1}}$ from the inversion trajectory. After the inner optimization completes for timestep $t_i$, a final backward Euler update is applied using the optimized unconditional embedding to obtain $z_{t_{i+1}}$. Repeating this process yields a sequence of unconditional embeddings $\{ e^{(i)}_{\mathrm{uncond}} \}$ and the Euler inverted noise latent $z_1^{\mathrm{ref}}$.

For editing, we reuse the recovered noise latent and the optimized unconditional embeddings, but replace the conditional embedding with that of the target edit prompt. Sampling with these components produces an edited asset.

\begin{algorithm}[h]
\caption{Euler Inversion and Null Text Optimization}
\label{alg:nullinv}

\KwIn{shape $X$, edit prompt $\mathcal{P}_{\text{edit}}$, steps $K$}
\KwOut{edited shape $\hat{X}$}

\textbf{// Preprocessing}
Voxelize and encode $X$ to obtain initial latent $z_{t_0}$ with $t_0 = 0$\;

\vspace{3pt}
\textbf{Inversion (cond = empty, $t_0=0 \rightarrow t_K=1$):}
\For{$i=0$ \KwTo $K-1$}{
    $v = v_\theta(z_{t_i}, t_i, \emptyset)$\;
    $z_{t_{i+1}} = z_{t_i} + (t_{i+1}-t_i)\,v$\;
    Save $z_{t_{i+1}}^{\mathrm{ref}}$\;
}
Store Euler inverted noise latent $z_{t_K}$\;

\vspace{3pt}
\textbf{Null Optimization (sampling, $t_0=1 \rightarrow t_K=0$):}
Set $z_{t_0} \leftarrow z^{\mathrm{ref}}_{t_K}$\;
\For{$i=0$ \KwTo $K-1$}{
    Initialize unconditional embedding $e^{(i)}_{\mathrm{uncond}}$\;
    \For{$j=1$ \KwTo num\_inner\_steps}{
        $v_{\mathrm{cond}} = v_\theta(z_{t_i}, t_i, \emptyset)$\;
        $v_{\mathrm{uncond}} = v_\theta(z_{t_i}, t_i, e^{(i)}_{\mathrm{uncond}})$\;
        $v = v_{\mathrm{cond}} + \omega (v_{\mathrm{cond}} - v_{\mathrm{uncond}})$\;
        $z^{\mathrm{pred}}_{t_{i+1}} = z_{t_i} - (t_i - t_{i+1})\, v$\;
        $\mathcal{L} = \|z^{\mathrm{pred}}_{t_{i+1}} - z^{\mathrm{ref}}_{t_{i+1}}\|$\;
        Update $e^{(i)}_{\mathrm{uncond}}$ using Adam on $\nabla\mathcal{L}$\;
    }
    \textbf{// final update for this timestep}
    $z_{t_{i+1}} = z_{t_i} - (t_i - t_{i+1})\, v$\;
}

\vspace{3pt}
\textbf{Editing:}
Generate $\hat{X}$ using the recovered noise latent $z_{t_K}$,
the sequence $\{ e^{(i)}_{\mathrm{uncond}} \}$,
and the edit prompt embedding for $\mathcal{P}_{\text{edit}}$\;

\end{algorithm}

\section{Evaluation of Geometric Expressivity}

In section~\ref{subsec:geo-expressivity}, we discuss the imbalance between language and geometric diversity. Indeed using an approximate prompt is detrimental to the inversion performance, both for standard Euler inversion and NTI. We present the qualitative results supporting this claim in Table~\ref{tab:eval_reconstruction}.

\begin{table}[htpb!]
  \centering
  \caption{Quantitative evaluation of reconstruction of DT4D shapes using different inversion methods. When using an approximate prompt $\Tilde{\mathcal{C}}$, the L1 and LPIPS score are notably higher. Even after null text optimization, the approximate prompt fails to faithfully reconstruct the original shape (\textit{third column}). The optimal combination is to use null text optimization and use an empty prompt during inversion.}
  \resizebox{0.65\columnwidth}{!}{
    \begin{tabular}{lllll}
    \toprule
    & Euler + $\Tilde{\mathcal{C}}$ & NTI + $\Tilde{\mathcal{C}} $ \cite{nti} &  Euler + $\emptyset$ & NTI+ $\emptyset$ \\
    \midrule
    L1    $\downarrow$ &   72.06 $\pm$ 62.95  &   15.57 $\pm$ 8.81    &   1.99 $\pm$ 1.60   &   \textbf{1.55 $\pm$ 1.46}   \\
    \midrule
    LPIPS $\downarrow$  &   0.34 $\pm$ 0.08    &   0.28 $\pm$ 0.08   &    0.25 $\pm$ 0.05   &   \textbf{0.24 $\pm$ 0.05}  \\ 
    \bottomrule
    \end{tabular}
}
\label{tab:eval_reconstruction}
\end{table}

\section{Texture Editing with TRELLIS}
\label{sec:supp_comp_trellis}

\begin{figure*}[t!]
    \centering
    \includegraphics[width=\textwidth]{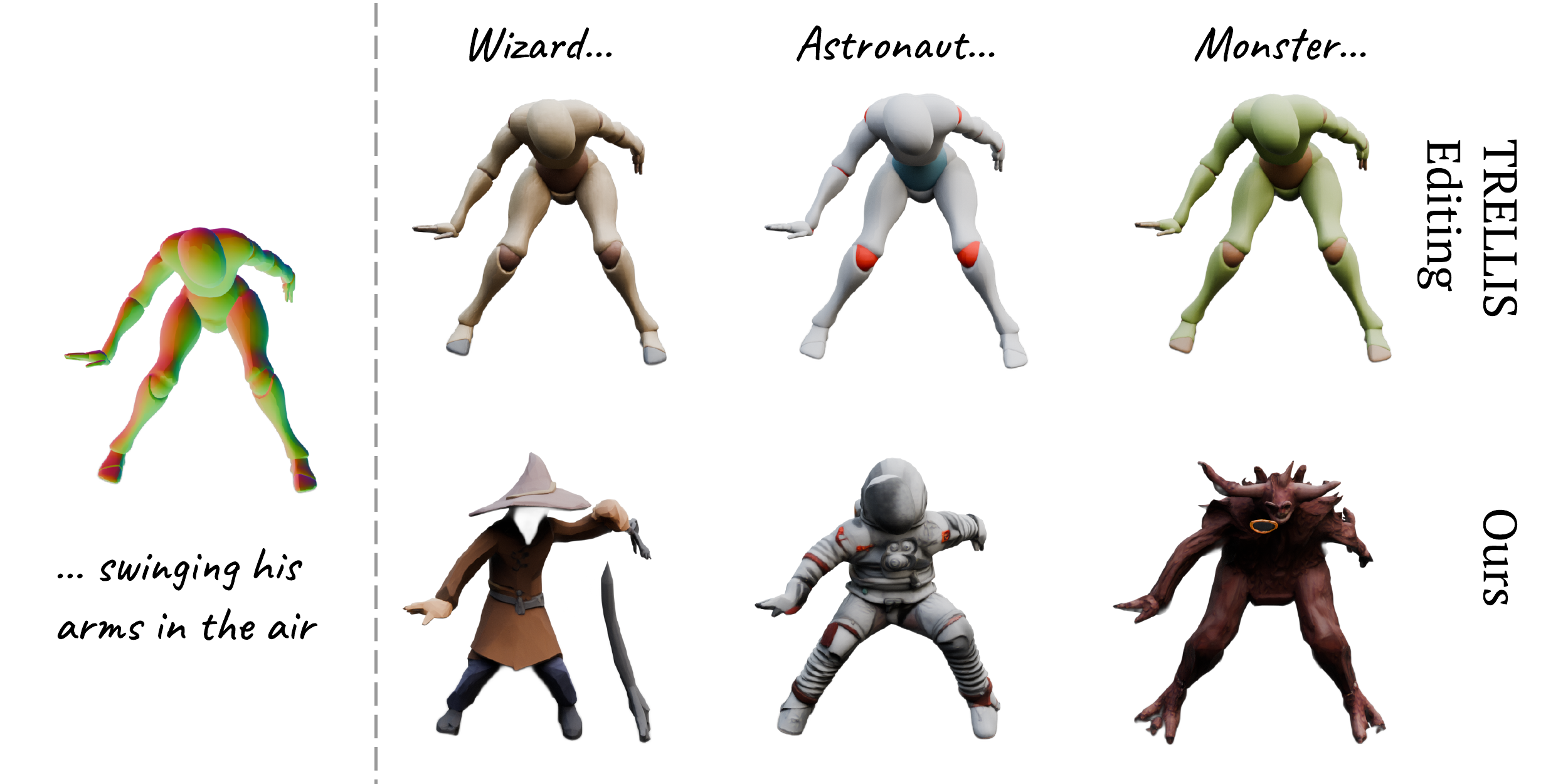}
    \caption{When using the editing method proposed by TRELLIS, the texture varies according to the edit prompt, but the coarse structure of the original asset remains unchanged. On the contrary, our method also modifies the overall geometry overall sample to better reflect the edit prompt.}
    \label{fig:trellis_editing}
\end{figure*}

The original TRELLIS paper proposes an editing method that retains the coarse structure and reruns only the second-stage $\mathcal{G}_L$ for refinement using a new prompt $\mathcal{P}$. While this approach can change appearance or texture, it offers limited control over geometry: the coarse structure largely determines the final shape, and only minor geometric details can be modified in the second stage. This behaviour is illustrated in Fig.~\ref{fig:trellis_editing}.




\section{Dataset Preparation}
\label{sec:supp_data_preparation}

\subsection{Shape editing of DT4D shapes}

To evaluate edits on DT4D~\cite{li2021dt4d}, we selected 13 sequences from 13 different characters, and sampled 16 random frames from each sequence. 
\begin{itemize}
    \item \texttt{moose45D\_Attack1}
    \item \texttt{raccoonVGG\_Run2}
    \item \texttt{chickenDC\_act3}
    \item \texttt{huskydog3T\_Actions2}
    \item \texttt{tigerD8H\_Actions5}
    \item \texttt{bearVGG\_Actions3}
    \item \texttt{Xbot\_Dancing}
    \item \texttt{knight\_JumpOver}
    \item \texttt{demon\_SalsaDancing}
    \item \texttt{malcolm\_WaveHipHopDance}
    \item \texttt{Maria\_Angry}
    \item \texttt{boss\_DrunkWalk},
    \item \texttt{alien\_StylishFlip}
    \item \texttt{pumpkinhulk\_HipHopDancing}
\end{itemize}

The precise edit prompts are in the supplementary material called ``dt4d\_\allowbreak edit\_\allowbreak prompts.json''. Note that in this json file, the ``source prompt'' are not what is used in our editing. It is included since we use the same JSON for generating edits with VoxHammer~\cite{li2025voxhammer}. The ``latent'' defines the original shape we meant to edit. 

\subsection{TRELLIS Generated Samples}

For both Sec.~\ref{subsec:language-diversity} and Sec.~\ref{subsec:ood-prompts}, we prompted TRELLIS using different prompts. We attach the prompts in the supplementary material as well, named ``sink\_\allowbreak experiment\_\allowbreak prompts.json'' and ``velocity\_\allowbreak experiment\_\allowbreak prompts\_\allowbreak trellis.txt''. The sampling parameters are the same as our editing experiments. Specifically, we use 50 sampling steps, with a CFG strength of 5 over the $t=0$ to $t=1$ interval.

\section{Additional Editing Results}
\label{sec:supp_editing_additional}
\begin{figure*}[t!]
    \centering
    \begin{subfigure}{0.24\textwidth}
        \centering
        \includegraphics[width=\linewidth]{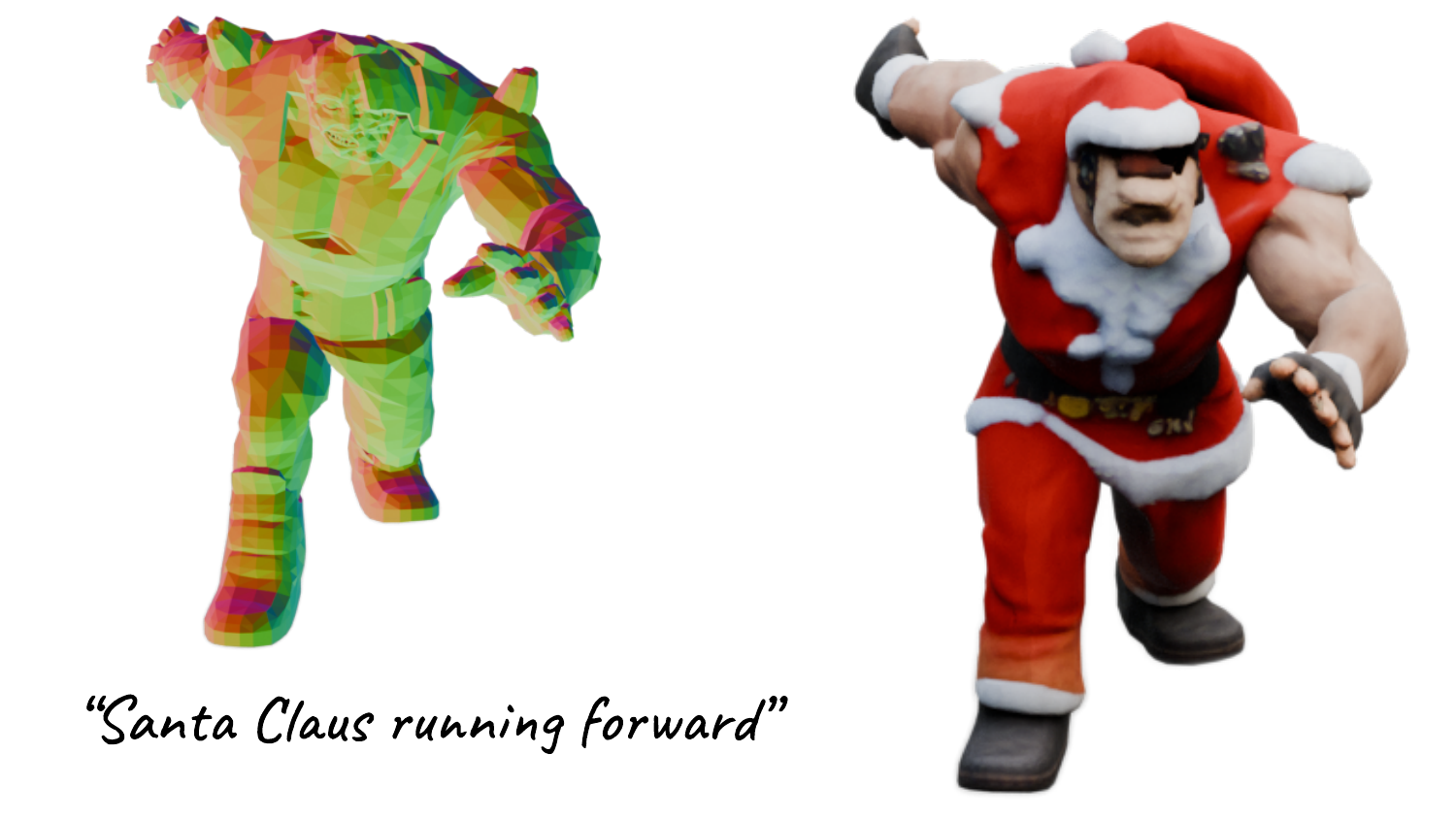}
    \end{subfigure}
    \begin{subfigure}{0.24\textwidth}
        \centering
        \includegraphics[width=\linewidth]{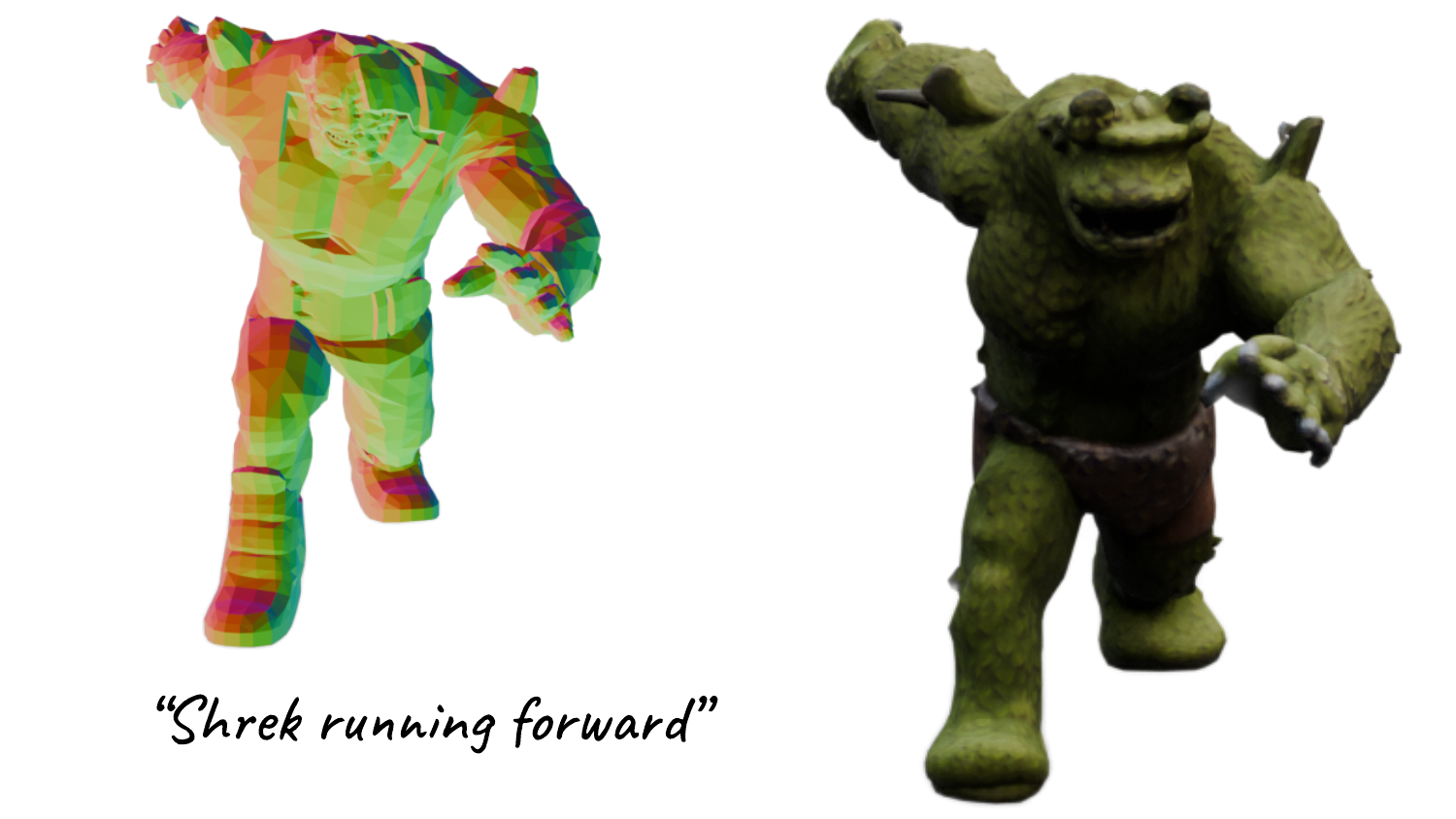}
    \end{subfigure}
    \begin{subfigure}{0.24\textwidth}
        \centering
        \includegraphics[width=\linewidth]{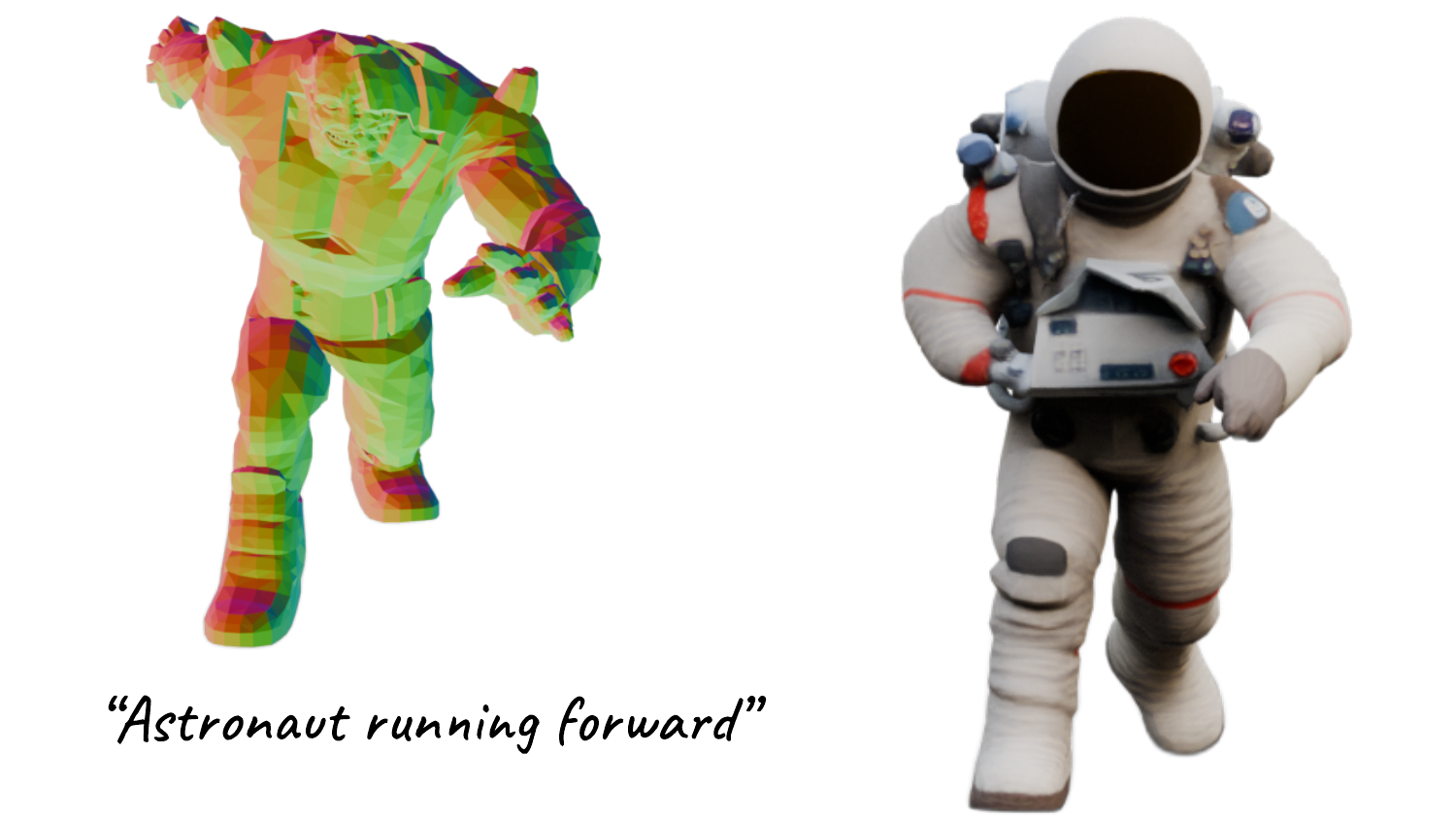}
    \end{subfigure}
    \begin{subfigure}{0.24\textwidth}
        \centering
        \includegraphics[width=\linewidth]{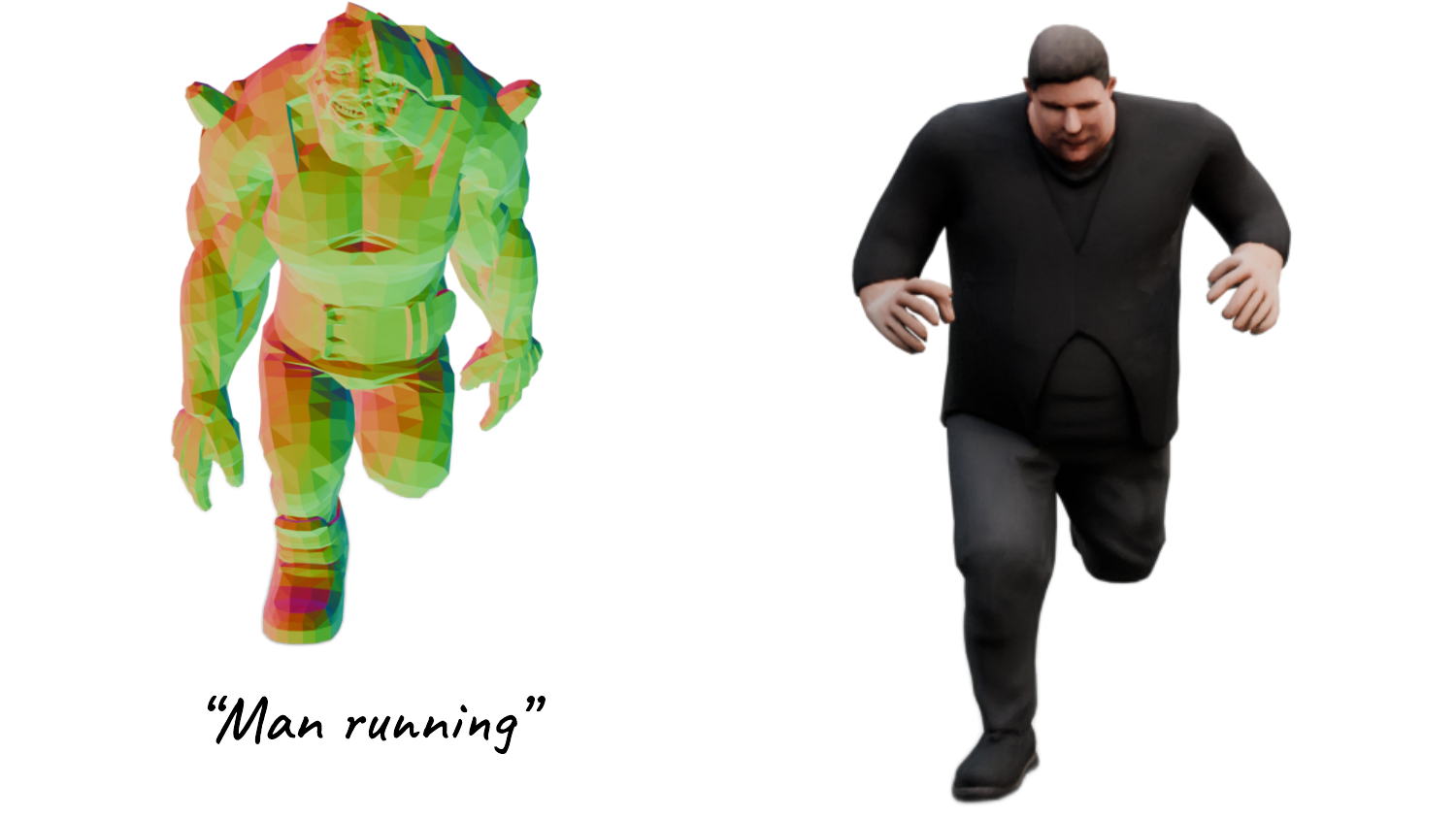}
    \end{subfigure}

    \begin{subfigure}{0.24\textwidth}
        \centering
        \includegraphics[width=\linewidth]{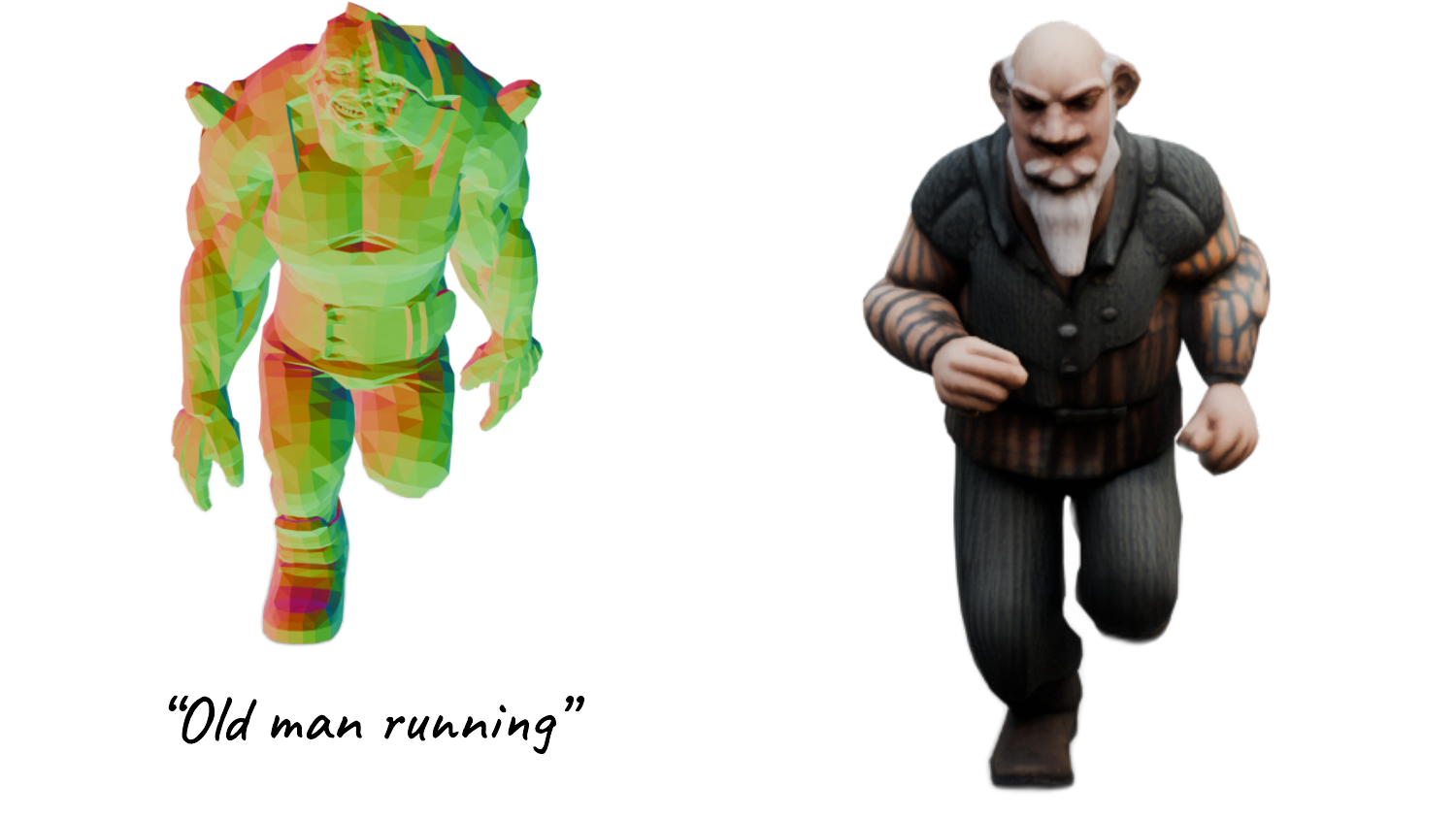}
    \end{subfigure}
    \begin{subfigure}{0.24\textwidth}
        \centering
        \includegraphics[width=\linewidth]{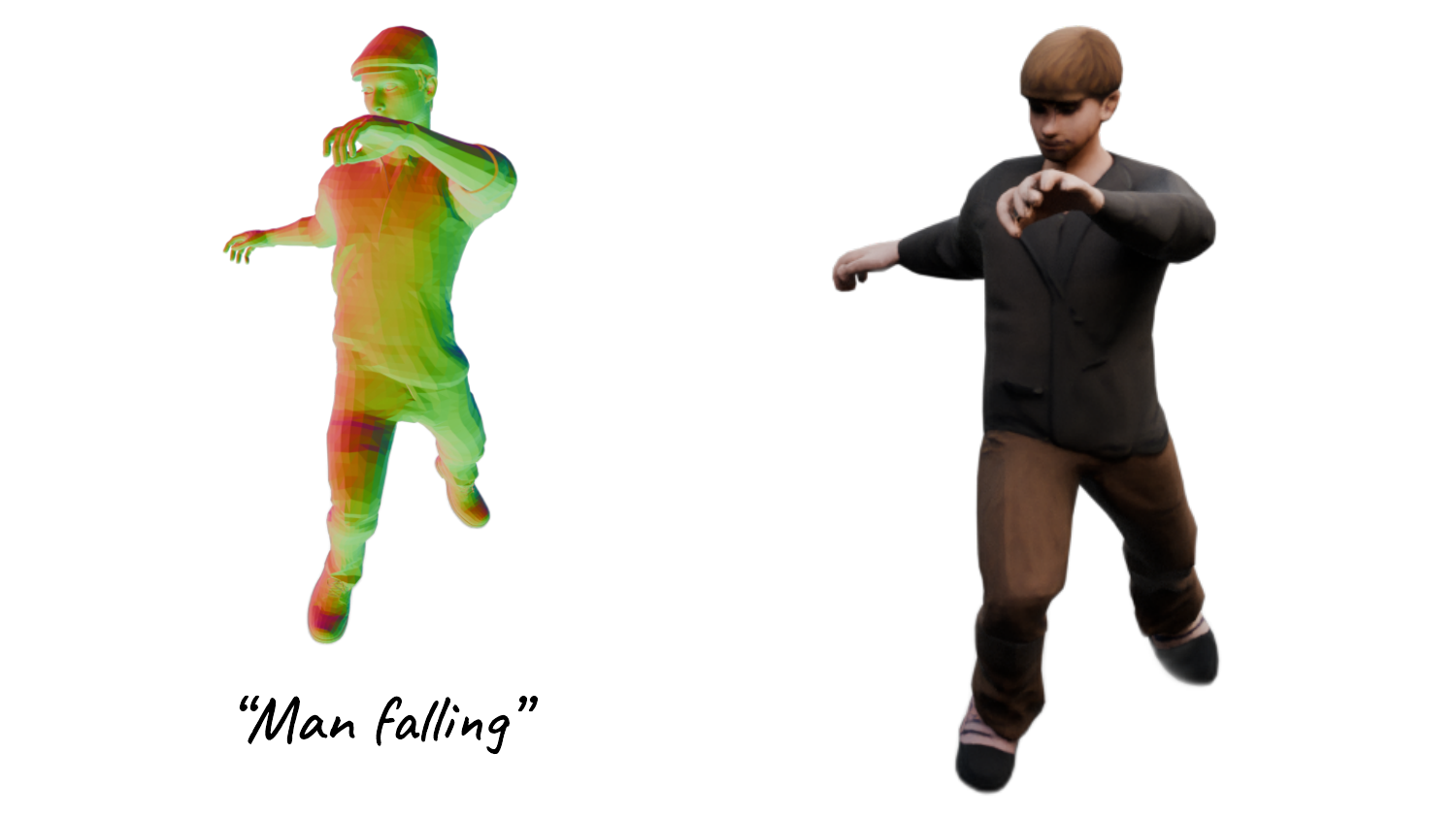}
    \end{subfigure}
    \begin{subfigure}{0.24\textwidth}
        \centering
        \includegraphics[width=\linewidth]{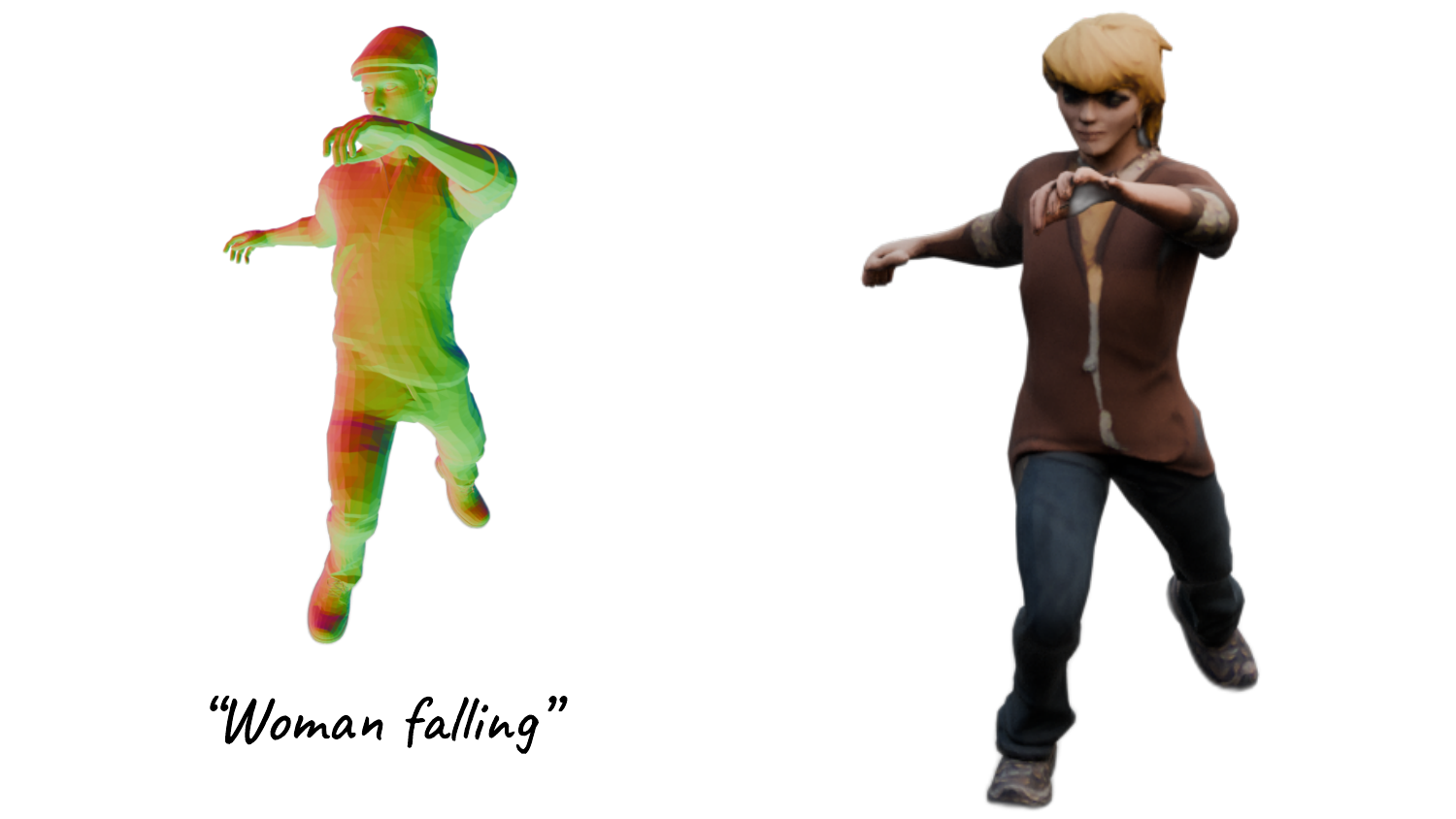}
    \end{subfigure}
    \begin{subfigure}{0.24\textwidth}
        \centering
            \includegraphics[width=\linewidth]{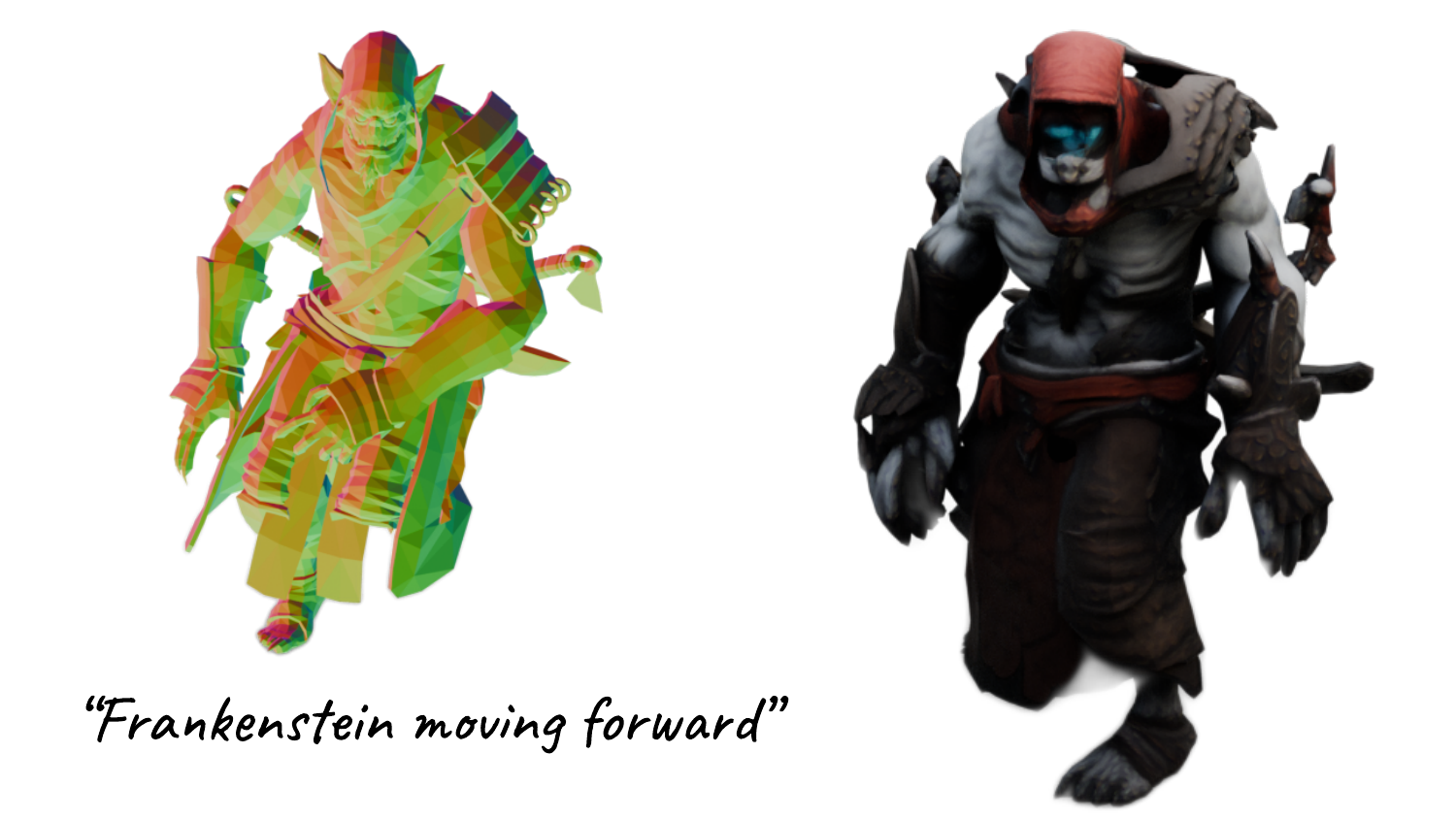}
    \end{subfigure}

    \caption{Additional edit examples. The input shape and the edit prompt $\mathcal{P}$ are shown on the right side of each sub-image}
    \label{fig:more_edits}
\end{figure*}

\begin{figure*}[t!]
    \centering
    \includegraphics[width=\textwidth]{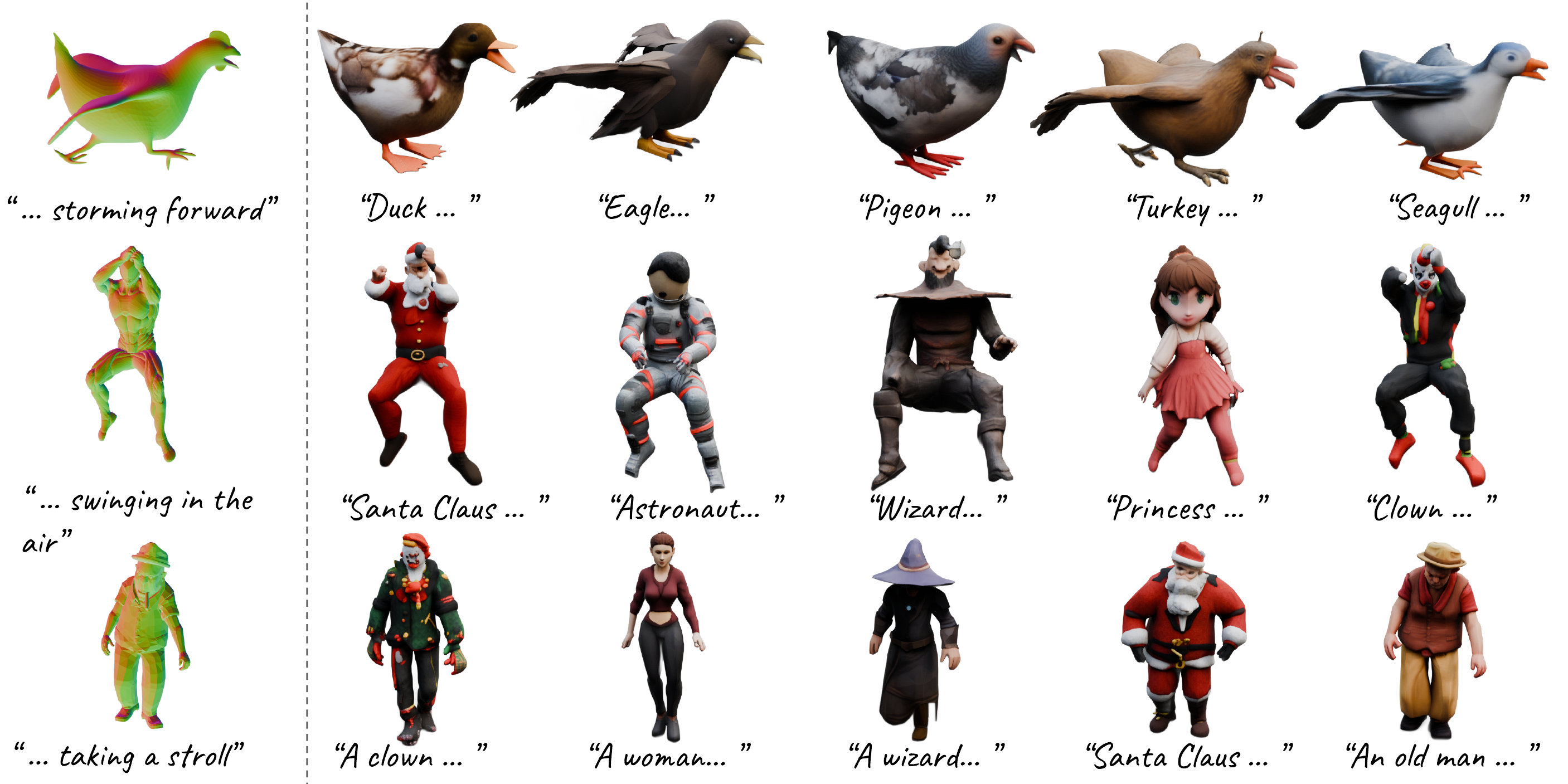}
    \caption{
    Additional editing examples demonstrating the versatility of our approach. For each row, the leftmost column shows the source shape inverted using our method, along with the shared prompt postfix $\mathcal{P}$ used across all edits. The remaining columns display edited results obtained by prepending different character descriptions to $\mathcal{P}$. Our method successfully transforms the source into diverse characters while consistently preserving the original pose and global structure across all variations.}
    \label{fig:editing_many_examples}
\end{figure*}

Additional editing results are shown in Fig.~\ref{fig:more_edits} and  Fig.~\ref{fig:editing_many_examples}.

\section{Inversion with Image Generative Models}
\label{sec:supp_inversion_image}

\begin{figure*}[t!]
\centering
\scalebox{0.70}{ 
\begin{minipage}{\textwidth}
    \centering
    \begin{subfigure}{0.24\textwidth}
        \centering
        \includegraphics[width=\linewidth]{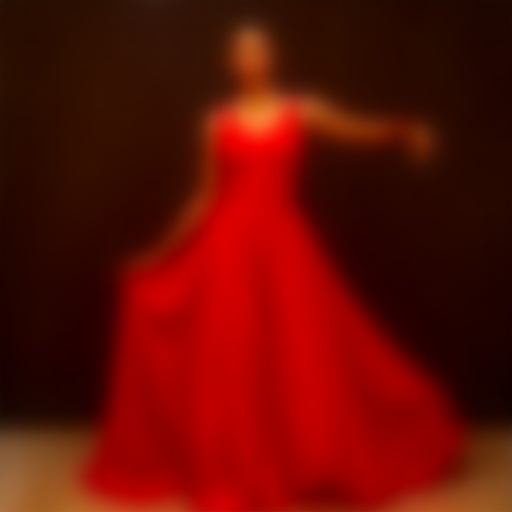}
    \end{subfigure}
    \begin{subfigure}{0.24\textwidth}
        \centering
        \includegraphics[width=\linewidth]{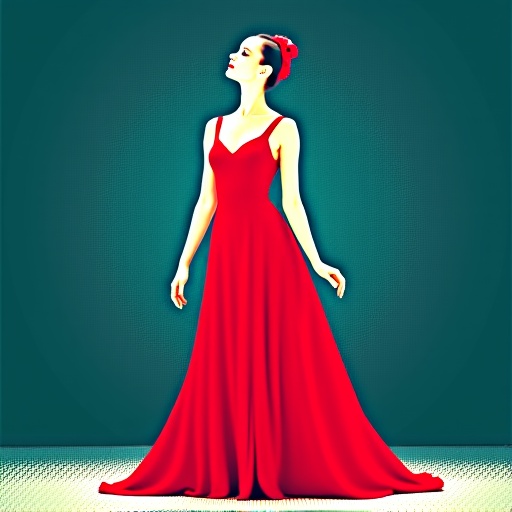}
    \end{subfigure}
    \begin{subfigure}{0.24\textwidth}
        \centering
        \includegraphics[width=\linewidth]{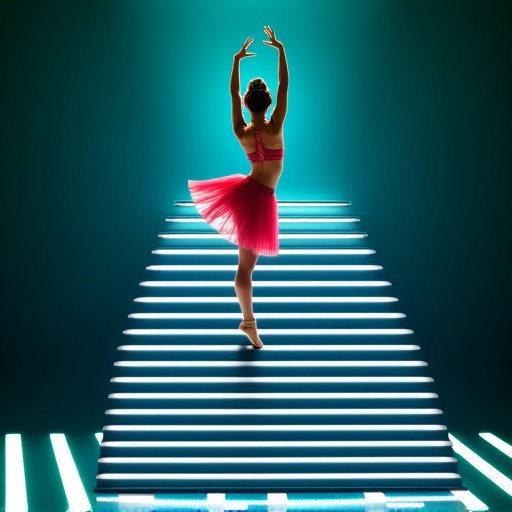}
    \end{subfigure}
    \begin{subfigure}{0.24\textwidth}
        \centering
        \includegraphics[width=\linewidth]{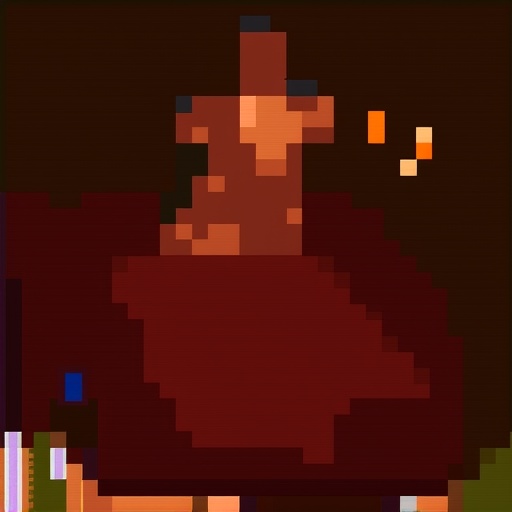}
    \end{subfigure}

    \begin{subfigure}{0.24\textwidth}
        \centering
        \includegraphics[width=\linewidth]{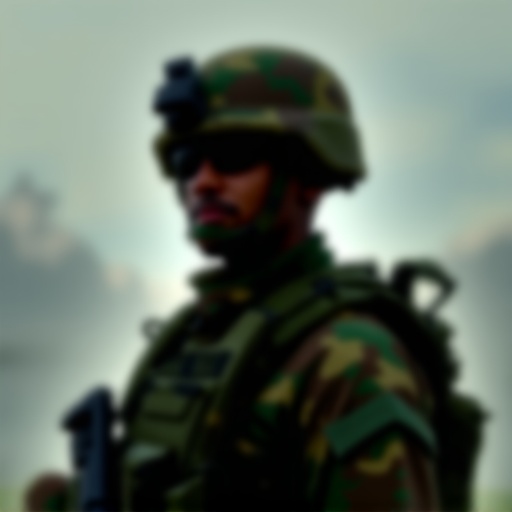}
    \end{subfigure}
    \begin{subfigure}{0.24\textwidth}
        \centering
        \includegraphics[width=\linewidth]{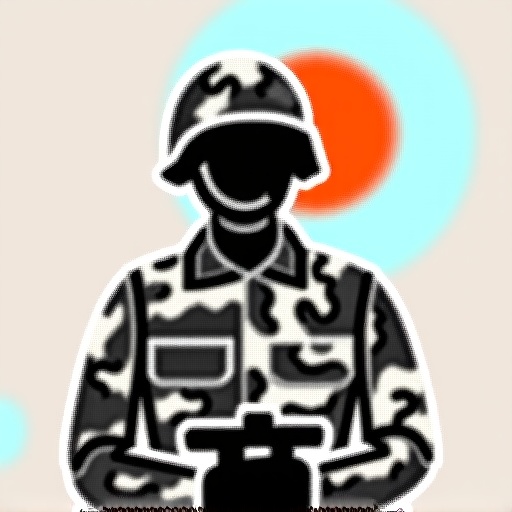}
    \end{subfigure}
    \begin{subfigure}{0.24\textwidth}
        \centering
        \includegraphics[width=\linewidth]{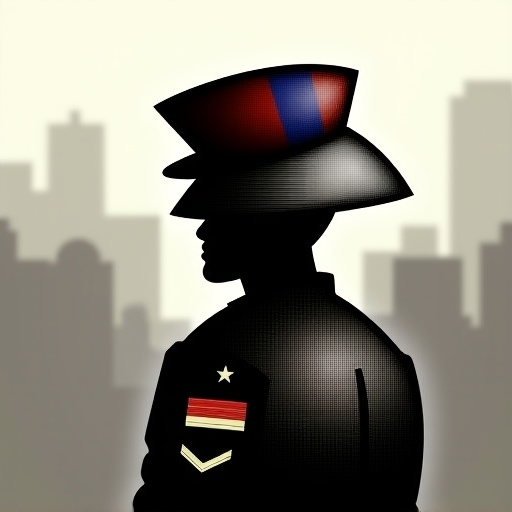}
    \end{subfigure}
    \begin{subfigure}{0.24\textwidth}
        \centering
        \includegraphics[width=\linewidth]{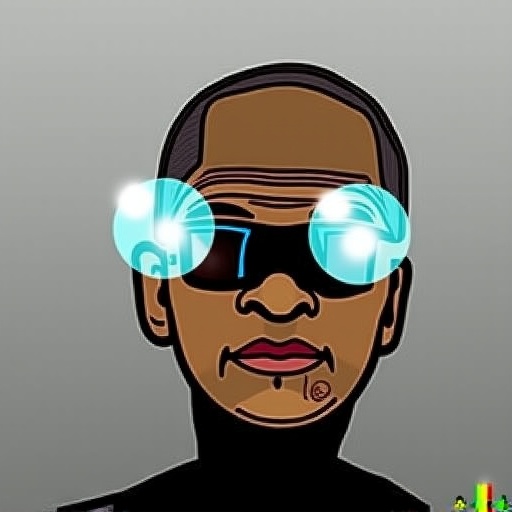}
    \end{subfigure}

    \begin{subfigure}{0.24\textwidth}
        \centering
        \includegraphics[width=\linewidth]{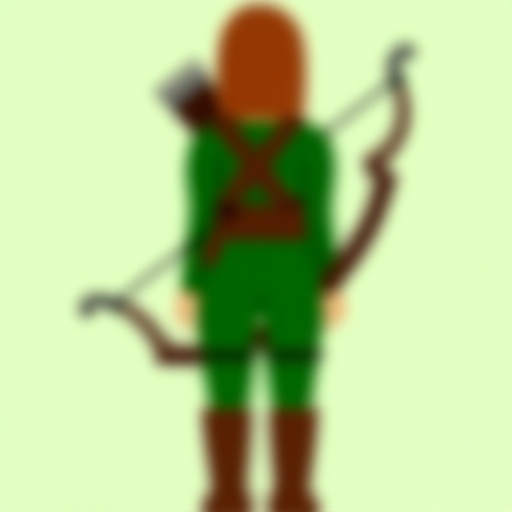}
    \end{subfigure}
    \begin{subfigure}{0.24\textwidth}
        \centering
        \includegraphics[width=\linewidth]{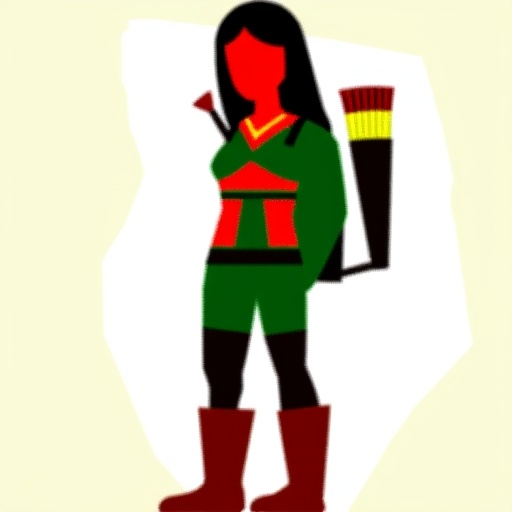}
    \end{subfigure}
    \begin{subfigure}{0.24\textwidth}
        \centering
        \includegraphics[width=\linewidth]{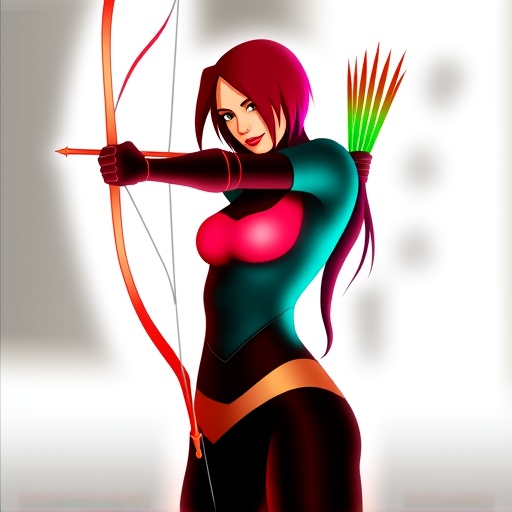}
    \end{subfigure}
    \begin{subfigure}{0.24\textwidth}
        \centering
        \includegraphics[width=\linewidth]{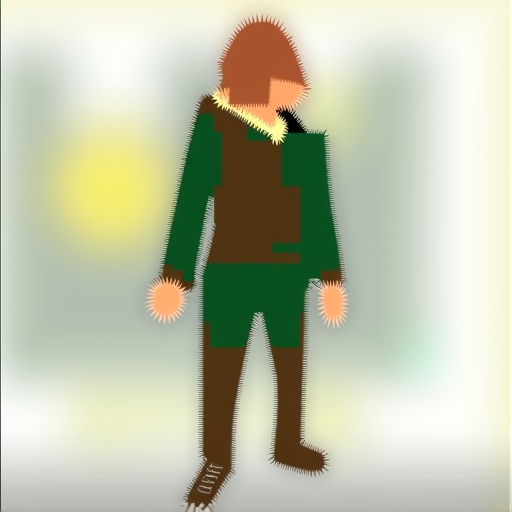}
    \end{subfigure}
    \vspace{2pt}
    \begin{subfigure}{0.24\textwidth}
        \centering
        \caption*{Source Image}
    \end{subfigure}
    \begin{subfigure}{0.24\textwidth}
        \centering
        \caption*{Using true prompt}
    \end{subfigure}
    \begin{subfigure}{0.24\textwidth}
        \centering
        \caption*{Using approximate prompt}
    \end{subfigure}
    \begin{subfigure}{0.24\textwidth}
        \centering
        \caption*{Using an empty prompt}
    \end{subfigure}
\end{minipage}
}
\caption{Visual comparison of inversion results on FLUX-generated images using the ground-truth prompt, an approximate prompt, and an empty prompt. Even when using an empty prompt for both inversion and resampling, the overall visual structure of the image is largely recovered.}
\label{fig:flux_experiment}
\end{figure*}

\begin{figure*}[t!]
\centering
\scalebox{0.70}{
\begin{minipage}{\textwidth}
    \centering
    \begin{subfigure}{0.24\textwidth}
        \centering
        \includegraphics[width=\linewidth]{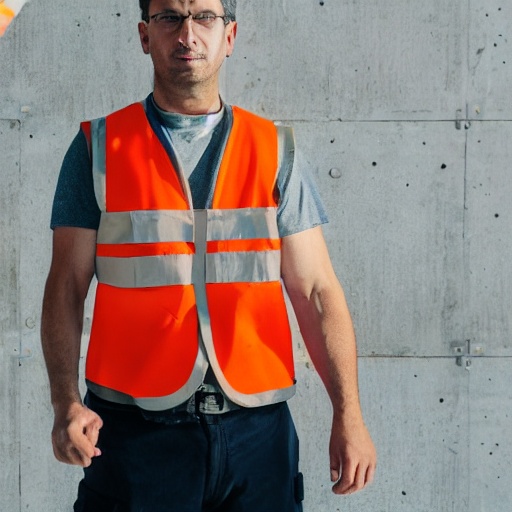}
    \end{subfigure}
    \begin{subfigure}{0.24\textwidth}
        \centering
        \includegraphics[width=\linewidth]{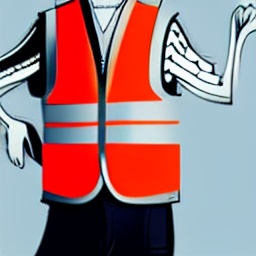}
    \end{subfigure}
    \begin{subfigure}{0.24\textwidth}
        \centering
        \includegraphics[width=\linewidth]{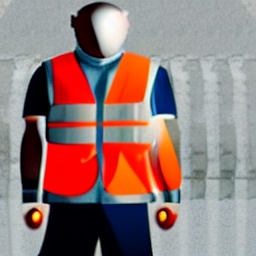}
    \end{subfigure}
    \begin{subfigure}{0.24\textwidth}
        \centering
        \includegraphics[width=\linewidth]{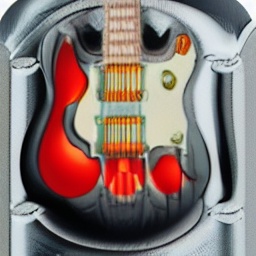}
    \end{subfigure}

    \begin{subfigure}{0.24\textwidth}
        \centering
        \includegraphics[width=\linewidth]{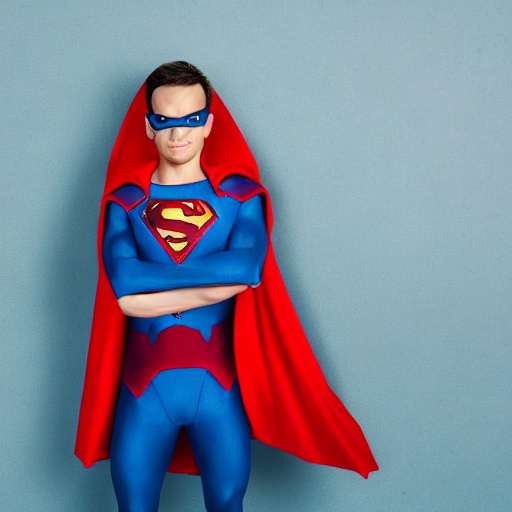}
    \end{subfigure}
    \begin{subfigure}{0.24\textwidth}
        \centering
        \includegraphics[width=\linewidth]{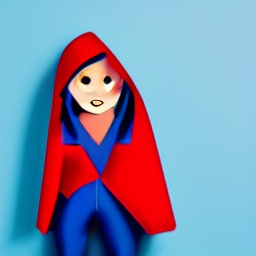}
    \end{subfigure}
    \begin{subfigure}{0.24\textwidth}
        \centering
        \includegraphics[width=\linewidth]{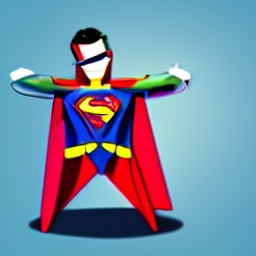}
    \end{subfigure}
    \begin{subfigure}{0.24\textwidth}
        \centering
        \includegraphics[width=\linewidth]{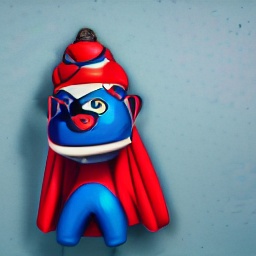}
    \end{subfigure}

    \begin{subfigure}{0.24\textwidth}
        \centering
        \includegraphics[width=\linewidth]{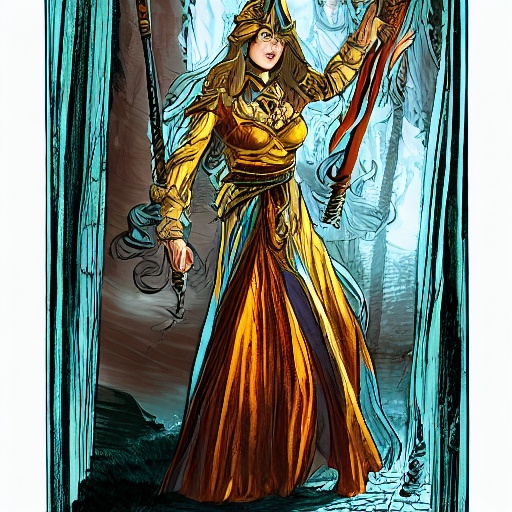}
    \end{subfigure}
    \begin{subfigure}{0.24\textwidth}
        \centering
        \includegraphics[width=\linewidth]{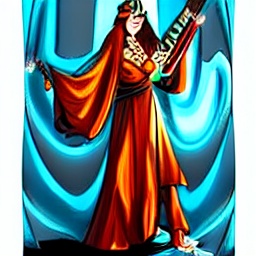}
    \end{subfigure}
    \begin{subfigure}{0.24\textwidth}
        \centering
        \includegraphics[width=\linewidth]{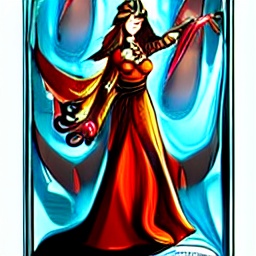}
    \end{subfigure}
    \begin{subfigure}{0.24\textwidth}
        \centering
        \includegraphics[width=\linewidth]{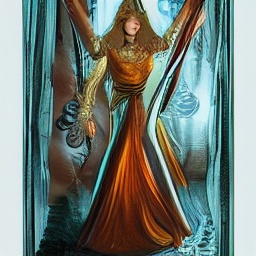}
    \end{subfigure}
    \vspace{2pt}
    \begin{subfigure}{0.24\textwidth}
        \centering
        \caption*{Source Image}
    \end{subfigure}
    \begin{subfigure}{0.24\textwidth}
        \centering
        \caption*{Using true prompt}
    \end{subfigure}
    \begin{subfigure}{0.24\textwidth}
        \centering
        \caption*{Using approximate prompt}
    \end{subfigure}
    \begin{subfigure}{0.24\textwidth}
        \centering
        \caption*{Using an empty prompt}
    \end{subfigure}
\end{minipage}
}
    \caption{Visual comparison of inversion results on SD-v1.4-generated images using the ground-truth prompt, an approximate prompt, and an empty prompt. All three prompt types yield reconstructions of similar quality relative to the source image. }
    \label{fig:sd_experiment}
\end{figure*}

In Sec.~\ref{sec:expressivity}, we noted that inversion in image generative models is comparatively insensitive to the choice of prompt: different prompt–noise combinations tend to follow similar inversion trajectories and yield comparable reconstructions. Here, we provide qualitative evidence supporting this observation. In Fig.~\ref{fig:flux_experiment} and Fig.~\ref{fig:sd_experiment}, we compare reconstructions obtained using three prompt types for FLUX~\cite{labs2025flux1} and Stable Diffusion~\cite{rombach2022sd}. For FLUX, empty prompts do not perfectly reconstruct the target, but they recover the main visual structure of the image, and using an approximate or ground-truth prompt produces similarly faithful results. For Stable Diffusion~\cite{rombach2022sd}, the insensitivity to the prompt type is even more pronounced: all three prompt types lead to reasonable reconstructions. These qualitative trends are consistent with the quantitative PSNR and LPIPS measurements reported in Tab.~\ref{tab:prompt_type_influence}.

\begin{figure*}[t!]
    \centering
    \begin{subfigure}{0.20\textwidth}
        \centering
        \includegraphics[width=\linewidth]{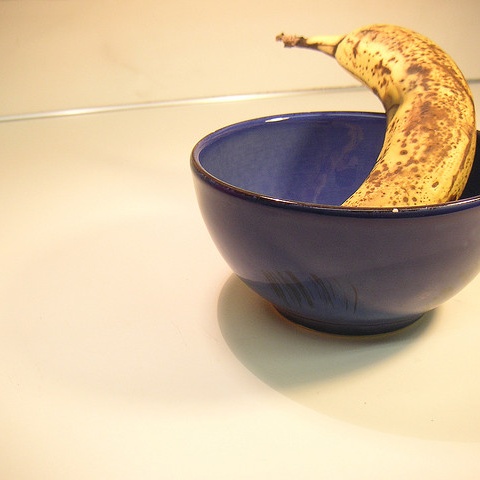}
    \end{subfigure}
    \begin{subfigure}{0.20\textwidth}
        \centering
        \includegraphics[width=\linewidth]{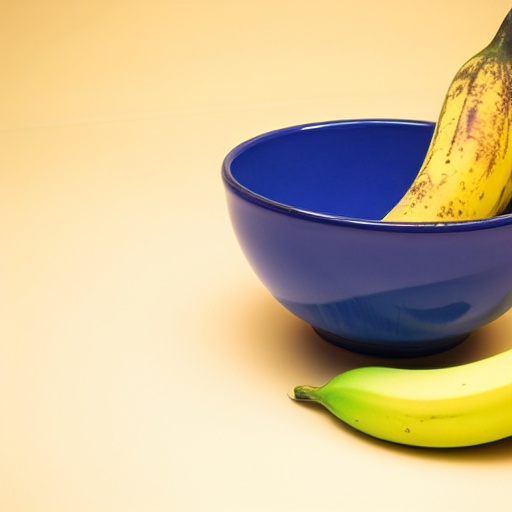}
    \end{subfigure}
    \begin{subfigure}{0.20\textwidth}
        \centering
        \includegraphics[width=\linewidth]{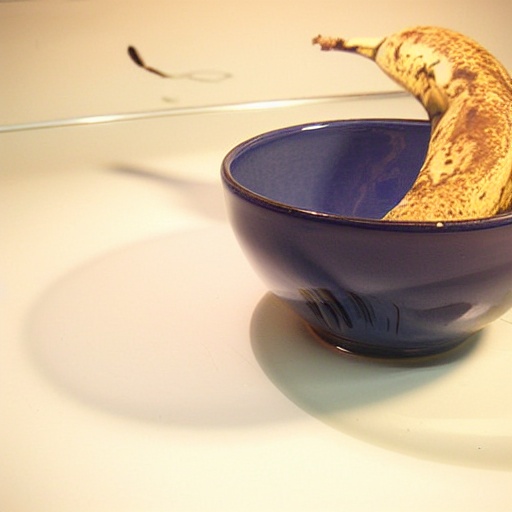}
    \end{subfigure}
    \vspace{1pt}
    
    \begin{subfigure}{0.20\textwidth}
        \centering
        \includegraphics[width=\linewidth]{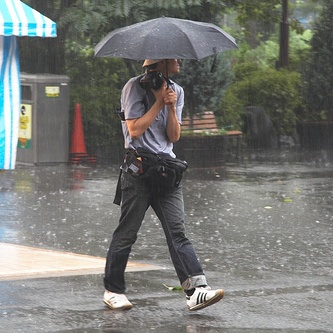}
    \end{subfigure}
    \begin{subfigure}{0.20\textwidth}
        \centering
        \includegraphics[width=\linewidth]{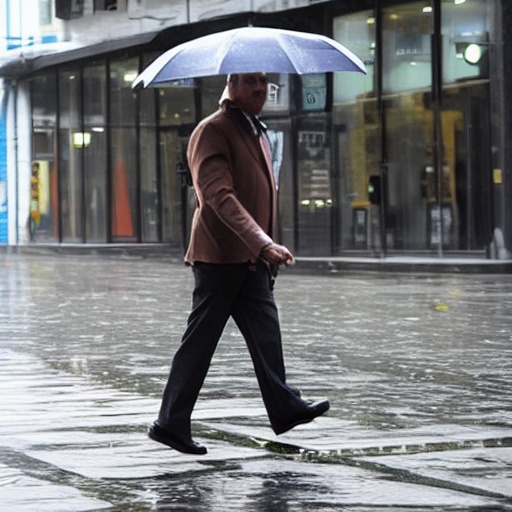}
    \end{subfigure}
    \begin{subfigure}{0.20\textwidth}
        \centering
        \includegraphics[width=\linewidth]{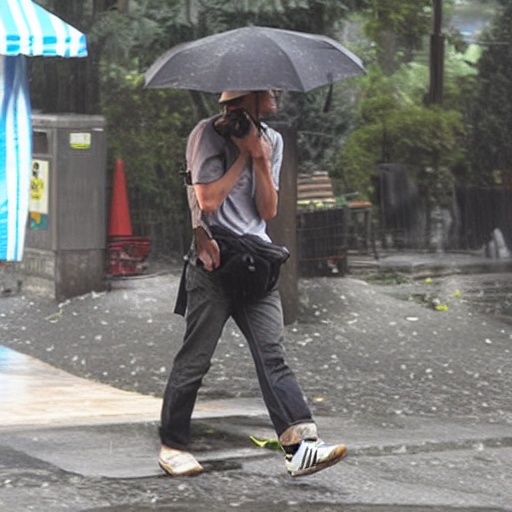}
    \end{subfigure}
    \vspace{1pt}

    \begin{subfigure}{0.20\textwidth}
        \centering
        \includegraphics[width=\linewidth]{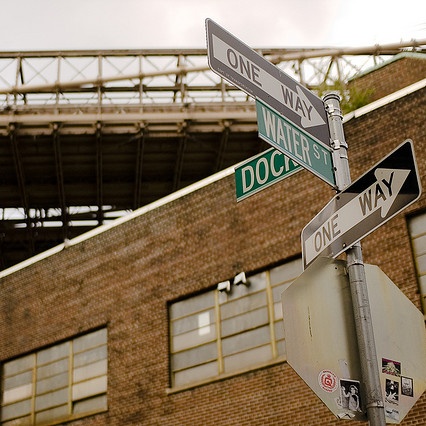}
    \end{subfigure}
    \begin{subfigure}{0.20\textwidth}
        \centering
        \includegraphics[width=\linewidth]{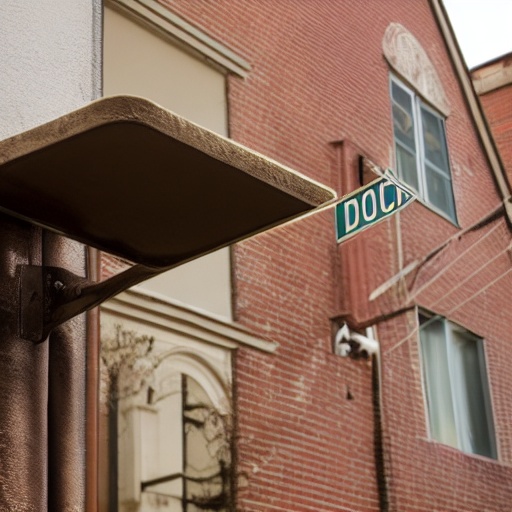}
    \end{subfigure}
    \begin{subfigure}{0.20\textwidth}
        \centering
        \includegraphics[width=\linewidth]{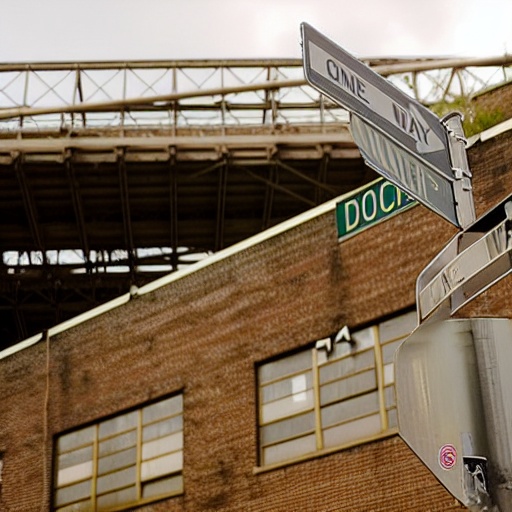}
    \end{subfigure}
        \vspace{1pt}

    \begin{subfigure}{0.20\textwidth}
        \centering
        \includegraphics[width=\linewidth]{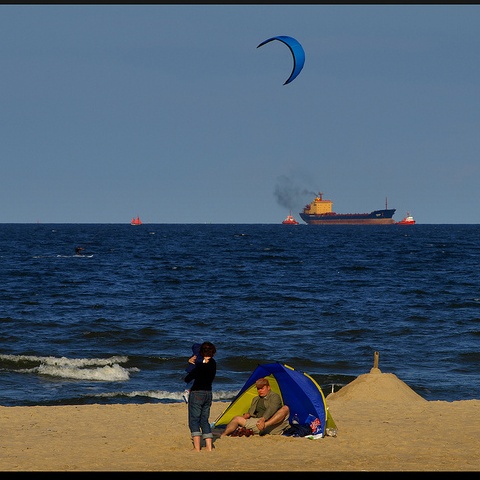}
    \end{subfigure}
    \begin{subfigure}{0.20\textwidth}
        \centering
        \includegraphics[width=\linewidth]{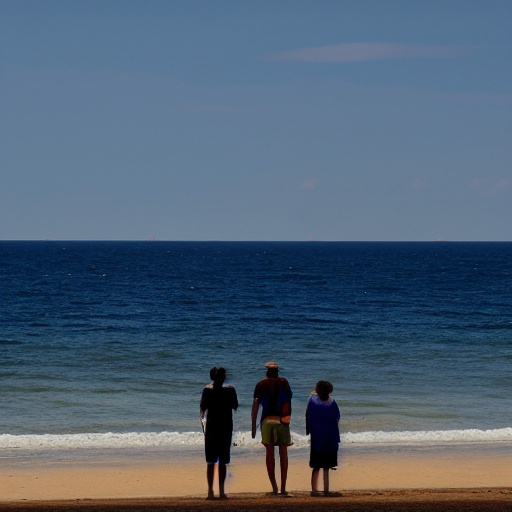}
    \end{subfigure}
    \begin{subfigure}{0.20\textwidth}
        \centering
        \includegraphics[width=\linewidth]{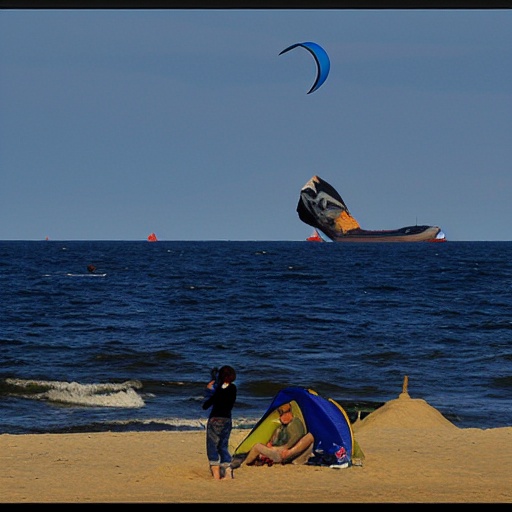}
    \end{subfigure}
        \vspace{1pt}

    \vspace{2pt}
    \begin{subfigure}{0.20\textwidth}
        \centering
        \caption*{Source Image}
    \end{subfigure}
    \begin{subfigure}{0.20\textwidth}
        \centering
        \caption*{Using true prompt}
    \end{subfigure}
    \begin{subfigure}{0.20\textwidth}
        \centering
        \caption*{Using empty prompt}
    \end{subfigure}

    \caption{Visual comparison of inversion results on SD-v1.4 using COCO images. Reconstructions using either the ground-truth caption or an empty prompt are of comparable quality, with empty prompts often producing higher-fidelity matches to the original image.}
    \label{fig:sd_experiment_coco}
\end{figure*}

Furthermore, for Stable Diffusion we conducted an additional experiment on COCO dataset~\cite{lin2015coco} and DiffusionDB~\cite{wang2023diffusiondb}, in which 100 randomly selected images were inverted and resampled using either their ground-truth caption or an empty prompt. Across both datasets, using an empty prompt often yielded improved reconstructions. Representative examples are shown in Fig.~\ref{fig:sd_experiment_coco}, with quantitative results provided in Tab.~\ref{tab:sd_on_prompt_type}.

\section{VLM-based Evaluation}
For VLM based evaluation of our text-driven edits, we used LLaVA-1.5-7B. We prompt the model to rate how well on the scale of 1-5 how well the edit prompt matches the resulting asset. We defined the rubric for scoring as follows:
\begin{enumerate}
\item Score 1 (No match): The resulting asset does not reflect the edit prompt at all. The edit completely failed.
\item Score 2 (Weak match): The asset barely reflects the prompt. It misses the main intent or major details.
\item Score 3 (Partial match): The asset reflects the core intent of the prompt, but has obvious errors, unintended artifacts, or missing elements.
\item Score 4 (Good match): The asset reflects almost everything in the prompt. The edit is successful with only minor flaws.
\item Score 5 (Perfect match): The asset exactly and fully reflects the edit prompt with no errors.
\end{enumerate}

For each model in our comparison, we report the average scores over the entire edit dataset.


\begin{figure*}[t!]
    \centering
    \includegraphics[width=0.85\textwidth]{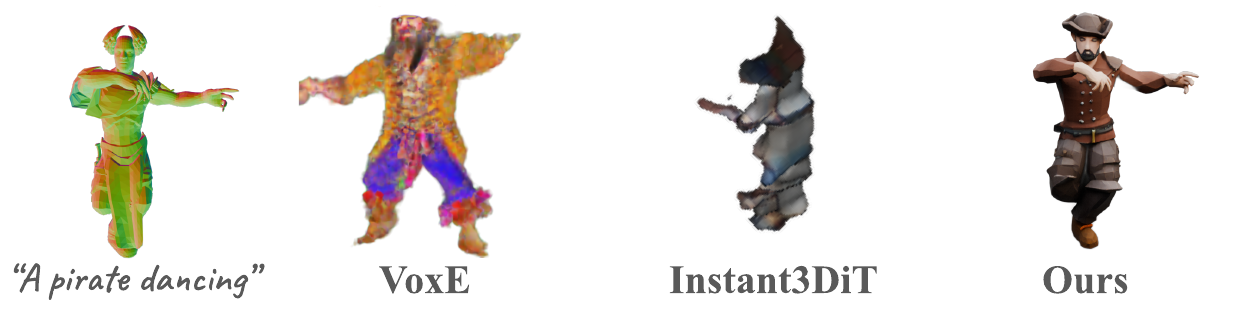}
    \caption{Comparison with Vox-E and Instant3DiT.}
    \label{fig:edit_fig_rebuttal}
\end{figure*}

\begin{table}[tb!]
  \centering
  \caption{Reconstruction accuracy when inverting images using ground-truth prompts versus empty prompts. Across both COCO and DiffusionDB, empty prompts yield higher reconstruction quality, supporting our observation that using an empty prompt improves inversion results also holds for image generative models.}
  \label{tab:sd_on_prompt_type}
  \begin{tabular}{lcc}
  \toprule
  SD v1.4~\cite{rombach2022sd} & With GT Prompt & Without Prompt \\
  \midrule
  \multicolumn{3}{c}{\textbf{PSNR} $\uparrow$} \\
  COCO & 15.73 $\pm$ 2.45 & \textbf{22.79 $\pm$ 3.67}\\
  DiffusionDB & 19.19 $\pm$ 2.55 & \textbf{25.71 $\pm$ 3.91} \\
  \addlinespace
  \multicolumn{3}{c}{\textbf{LPIPS}~\cite{zhang2018lpips} $\downarrow$} \\
  COCO & 0.595 $\pm$ 0.097 & \textbf{0.507 $\pm$ 0.103}\\
  DiffusionDB & 0.448 $\pm$ 0.188 & \textbf{0.388 $\pm$ 0.228}\\
  \bottomrule
  \end{tabular}
\end{table}

\end{document}